\documentclass[letterpaper]{article} 
\usepackage{aaai2026}  
\usepackage{times}  
\usepackage{helvet}  
\usepackage{courier}  
\usepackage[hyphens]{url}  
\usepackage{graphicx} 
\usepackage{natbib}  
\usepackage{caption} 
\usepackage{algorithm}
\usepackage{algorithmic}

\usepackage{newfloat}
\usepackage{listings}
\DeclareCaptionStyle{ruled}{labelfont=normalfont,labelsep=colon,strut=off} 
\floatstyle{ruled}
\newfloat{listing}{tb}{lst}{}
\floatname{listing}{Listing}
\usepackage{subcaption}
\usepackage{amssymb}
\usepackage{algorithm}
\usepackage{algorithmic}
\usepackage{amsmath}
\usepackage{xspace}
\usepackage{siunitx}
\usepackage{makecell}
\usepackage{fancyvrb}
\usepackage{multirow}
\usepackage[dvipsnames]{xcolor} 
\usepackage{subcaption}
\usepackage{dsfont}
\usepackage[table]{xcolor}
\usepackage{amsthm}
\usepackage{thmtools}
\usepackage{multicol}
\usepackage{enumitem}
\usepackage{booktabs}
\usepackage{pifont}

\newcommand{\APPROACH}{Monte Carlo Query Synthesis\xspace}
\newcommand{\APPABBREV}{MCQS\xspace}
\newtheorem{definition}{Definition}
\newtheorem{theorem}{Theorem}
\newtheorem{lemma}{Lemma}
\usepackage{tcolorbox}
\tcbset{
  colback=gray!5,
  colframe=black,
  boxrule=0.5pt,
  arc=2pt,
  left=6pt,
  right=6pt,
  top=6pt,
  bottom=6pt
}
\newtcolorbox{capbox}[2][]{
  title=\textsc{#2},
  #1
}

\newcommand{\mM}{h}
\newcommand{\set}[1]{\{#1\}}
\newcommand{\tuple}[1]{\langle #1 \rangle}

\newcommand{\hset}{h^{\dagger}\xspace}
\newcommand{\hsettrue}{h^{\dagger\star}\xspace}

\title{Autonomous Assessment of Generalizability of AI Agent Capabilities}
\author{
    Daniel Bramblett\textsuperscript{\rm 1}, Rushang Karia\textsuperscript{\rm 1}, Adrian Ciotinga\textsuperscript{\rm 1}, 
    Pulkit Verma\textsuperscript{\rm 2},\\ \textbf{YooJung Choi}\textsuperscript{\rm 1}, \textbf{Siddharth Srivastava}\textsuperscript{\rm 1}\\
}
\affiliations{
     \textsuperscript{\rm 1}Arizona State University, AZ, USA\\
     \textsuperscript{\rm 2}Indian Institute of Technology Madras, India\\
\{drbrambl,rkaria,acioting,yj.choi,siddharths\}@asu.edu,
pulkitv@cse.iitm.ac.in
}

\usepackage{bibentry}

\begin{document}

\maketitle

\begin{abstract}
Safe deployment of black-box AI (BBAI) systems such as foundation model agents requires methods for evaluating their capabilities in novel settings. We define an agent’s capability as its ability to achieve a short term objective and formalize the problem of learning models that predict whether, with what effects, and under what conditions, an agent can perform a capability. We introduce \APPROACH{} (\APPABBREV), an active query-synthesis method for learning symbolic stochastic capability models of BBAIs. \APPABBREV{} models capabilities as conditional probability distributions over outcomes and formulates capability evaluation as an active learning problem over policies. We use Monte Carlo tree search to synthesize queries that maximally distinguish between extremal capability hypotheses: the lattice meet and join corresponding to the most pessimistic and optimistic models consistent with observed behavior. Executing these queries yields trajectories that prune inconsistent hypotheses. We prove soundness, completeness, and convergence properties under standard realizability and sampling assumptions. Experiments with multiple BBAI systems show that \APPABBREV{} learns accurate capability models more efficiently than baseline query strategies, enabling systematic characterization of agent capability boundaries with fewer interactions.
\end{abstract}

\section{Introduction}

\begin{figure*}[t]
\noindent\begin{minipage}{0.11\linewidth}
    \includegraphics[width=4\linewidth]{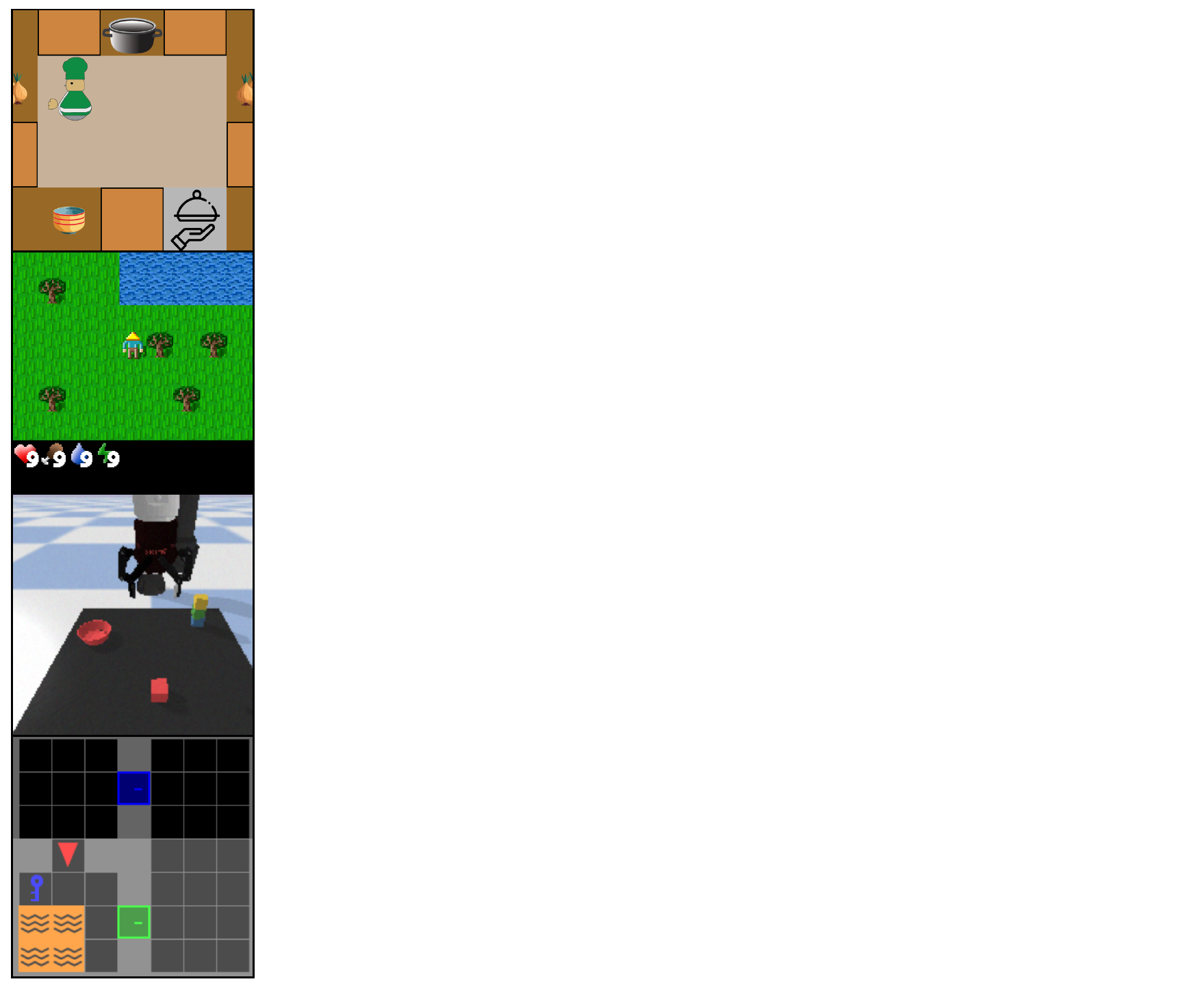}

    \subcaption{}
    \label{fig:agent_imgs}
\end{minipage} 
\begin{minipage}{0.8\linewidth}
\begin{small}
\begin{tabular}
{p{0.15\linewidth}p{0.19\linewidth}p{0.64\linewidth}}
\toprule
\textbf{Agent}& \textbf{Capability} & \textbf{Conditions \& Outcomes}\\
\midrule
\multirow{3}{\linewidth}{%
    GPT-5.4  on Minigrid
} & Pick up blue key (in SW) & From NW or SW, achieves intent with ${\sim}0.7$ probability; side effect: opens unnecessary  door in NW with ${\sim}0.5$ probability.\\
& Open blue door & From SW, always achieves intent when not holding key;  but probability of success drops to $0.8$ when holding key. \\
& Go to SW & From NW, achieves the intent with ${\sim}0.2$ probability when unnecessary door is closed, but only ${\sim}0.1$ when it is open; in the latter case, the agent ends up in SE.\\
\midrule
\multirow{2}{\linewidth}{%
    CSteve1  on Crafter
} & Move away from tree & Achieves intent by cutting down tree.\\
& Achieve drink water & In some scenarios moves away from water instead (probability ${\sim}0.7$).\\
\midrule
SayCan & Stack green on yellow & Achieves intent (${\sim}0.6$); fails by picking up entire tower (${\sim}0.4$); numerous error outcomes (unstack/knock-down).\\
\midrule
LAO* on Blocksworld & Place block A & Always prefers placing on C if clear; otherwise on B.\\
\bottomrule
\end{tabular}
\subcaption{}
\label{tab:capabilities}
\end{small}
\end{minipage}
\caption{\small (a)  Evaluated BBAIs and environments from top-down: Overcooked, Crafter, Saycan, and Minigrid; (b) some of the salient capabilities discovered by \APPABBREV{}. See Sec.\,\ref{sec:empirical_results} for details.} \label{fig:spoiler}
\end{figure*}

We consider the problem of efficiently learning interpretable capability models of black-box AI systems (BBAIs) capable of planning and learning. 
Modern BBAIs such as Large Language Models (LLMs), Vision-Language Models (VLMs), etc.\
accept high-level instructions and perform  long-horizon sequential decision-making~\citep{ha2023ai,liu2023visual,intelligence2025pi_}. However, it is difficult to predict what such BBAIs can and cannot do.
%
%
%
For instance, our experiments show that spurious correlations in training data can induce RL agents to exhibit ``superstitious'' behaviors, such as opening an unnecessary door when the goal is to retrieve a key. Such configuration-dependent effects complicate safe deployment and limit the reliable use of AI systems.
Prior work on the topic is largely restricted to agents with pre-determined capabilities that can be modeled in relatively simple modeling languages (e.g.\ QACE \citep{verma23neurips}). Another related direction of work, on action-model learning (Sec.\,\ref{sec:related_work}), learns models describing how an agent's primitive actions affect its environment.  In contrast, we focus on the relatively under-studied problem of learning \emph{capability models} that characterize the types of planning problems an agent can solve, the conditions under which it can solve them, and their possible outcomes. This is a challenging open problem because we cannot assume prior knowledge of a BBAI's internal representations and planning algorithms.
 Algorithms for action model learning do not address this setting and existing approaches for capability learning cannot be used in stochastic environments involving agents with non-stationary policies or evolving capabilities. However, solving this problem is essential for determining the limits of generalizability of planning behavior in popular BBAIs.  E.g., Fig.\,\ref{fig:spoiler} shows some of the discovered  agent capabilities and their surprising limitations found using the approach presented in this paper. Each capability may involve the execution of multiple actions.

We formalize capability discovery and learning as an active learning problem over a hypothesis class $\mathcal{H}$ of capability models. Each model $h\in\mathcal{H}$ induces a distribution over state-trajectories resulting from BBAI's execution of its capabilities. Unlike standard active learning, the query space here is the space of policies that map states to capabilities, which grows exponentially with the state space.

Our main contribution is a class of Monte Carlo Query Search (MCQS) algorithms for constructing queries for this policy-level active learning problem. The key challenge is the combinatorial structure of both the policy space and the hypothesis class. We address this via two observations. First, the set of hypotheses consistent with observed BBAI executions (state trajectories) constitutes a subset lattice. We show that it suffices to track its extremal elements, i.e., the join and meet of all possible consistent hypotheses ($h_\lor$ and $h_{\land}$ respectively), which can be computed efficiently without enumerating individual hypotheses. Second, inspired by Bell experiments~\citep{bell1964einstein}, we define discriminating policies as those that induce disparate trajectory distributions under the hypotheses $h_\lor$ and $h_{\land}$. Concretely, we measure a policy's ability to discriminate among the two competing hypotheses in the form of expected divergence between trajectory distributions induced by each hypothesis, and treat this divergence measure as a maximization objective for policy computation. Together, these ideas lead to a principled approach for computing policies that are highly informative.

We make minimal assumptions, only requiring access to a \emph{representation function} that maps environment states into a user-interpretable symbolic space. While a substantial body of work studies the problem of learning representation functions~\citep{shah2025from,konidaris2018skills,ahmetoglu2022deepsym,peng2024preference,james2020learning}, we focus on the complementary problem of learning capability models given a fixed representation. 

The main contributions of this work are: (1) a capability-modeling framework for BBAIs, with a formal learning objective (Sec.~\ref{sec:formal_frame}); (2) \APPABBREV{}, an approach for capability discovery and learning via informative query synthesis (Sec.~\ref{sec:mcqs}); (3) theoretical results showing that \APPABBREV{} converges in the limit (Sec.~\ref{sec:results}); and (4) empirical results demonstrating its scope and effectiveness across diverse agents and environments (Sec.~\ref{sec:results}).

\section{Formal Framework}
\label{sec:formal_frame}

\label{sec:problem_setting}

We evaluate BBAIs operating in a stochastic, fully observable environment $\mathcal{E}$ characterized by environment states $X$ and low-level actions $A$. We assume access to a simulator $\mathcal{S}_\mathcal{E}$ for $\mathcal{E}$ that supports standard functionality: resetting to an initial state, reverting to any previously encountered state $x \in X$, stepping the simulator with an action $a\in A$ to obtain the next state and outcome, and querying the set of available actions. We otherwise assume neither explicit knowledge of $\mathcal{E}$ nor additional functionality from $\mathcal{S}_\mathcal{E}$.

Environment states $X$ are often uninterpretable to users, motivating a high-level symbolic representation for expressing both states and capability models. We assume an interpretable symbolic state space $S$ defined over a vocabulary $V$ consisting of a set of predicate symbols $P$ and a set of objects $O$, where ground atoms are formed by applying predicates in $P$ to objects in $O$. 
We use a representation function $\alpha: X \rightarrow S$ where $\alpha(x)$ is the set of ground atoms that hold in $x$.
As noted earlier (Sec.\,\ref{sec:introduction}), several approaches are being developed for learning such representations, yet the problem of using them for capability model learning remains understudied and forms the focus of this paper.


\subsection{Capability Learning Problem}
\label{sec:ml_objective}

Intuitively, each capability corresponds to a latent intent-conditioned option~\citep{sutton1999between}. We consider intents $\mathcal{I}$, where each intent is a conjunction of ground atoms over $V$. We use ``goal'' for a user-assigned objective and ``intent'' for the agent's current short-term objective, possibly pursued as a step toward a goal. Each intent $i \in \mathcal{I}$ is associated with a latent option consisting of an initiation set, policy, and termination function. Let $\Delta(A)$ denote the set of probability distributions over $A$. Formally:

\begin{definition}
Let $\mathcal{A}$ be a BBAI and $\Omega$ be a set of options. A \emph{capability function} of $\mathcal{A}$ is a mapping $\kappa : \mathcal{I} \rightarrow \Omega$ such that $\kappa(i) = \langle S^i_0, \mu_i, \beta_i \rangle$, where $S^i_0 \subseteq S$ is the initiation set, $\mu_i : S \rightarrow \Delta(A)$ is a policy for achieving intent $i$, and $\beta_i : S \rightarrow \set{0,1}$ is the termination function.
\end{definition}

\begin{figure}
  \begin{center}
    \small
    \centering
    \begin{minipage}[t]{0.95\linewidth}
    \textbf{Capability Name:} $c_2$;
    \textbf{Intent:} $\textrm{clean}(l_1)$\\[4pt]
    \textbf{Conditional Effect $r_n$:}\\
    \quad\textbf{Condition:}
    $
    \textrm{charged}(\textrm{robot})
    \vee
    \textrm{at}(\textrm{charger},\textrm{robot})
    $
    

\textbf{Effects:}\\[-8pt]
\hspace*{1.5cm}$
\begin{array}{ll}
0.75: & \textrm{clean}(l_1) \wedge
        \textrm{at}(\textrm{charger},\textrm{robot})\\
0.25: & \neg \textrm{charged}(\textrm{robot})
\end{array}
$

\vspace{1mm}
\textbf{Conditional Effect $r_{n+1}$:}\\
    \quad\textbf{Condition:}
    $
    \neg\textrm{charged}(\textrm{robot})
    \wedge
    \neg\textrm{at}(\textrm{charger},\textrm{robot})
    $
    

    \textbf{Effects:}\\[-8pt]
    \hspace*{1.5cm}$
    \begin{array}{ll}
    0.10: & \textrm{at}(\textrm{charger},\textrm{robot})\wedge\textrm{charged}(\textrm{robot}) \\
    0.90: & \textrm{request\_help}(\textrm{user})
    \end{array}
    $
    \end{minipage}
    \caption{\small Simplified example of a capability model.}
    \label{fig:capability_ex}
  \end{center}
\end{figure}

We write the capability for intent $i$ as $c=\kappa(i)$. Let $C$ be the set of capabilities under consideration. Our objective is to learn a model for each $c\in C$ approximating its initiation set $S^i_0$ and the effects induced by executing $\mu_i$ from states in $S^i_0$ until termination under $\beta_i$. 


To be able to express models of real BBAIs, we model capabilities using conditional probabilistic effects, where conditions may include conjunctions and disjunctions, and outcomes are stochastic.
Fig.\,\ref{fig:capability_ex} shows a simplified example for a robot vacuum cleaner's capability for achieving the intent of cleaning a location.

We use grounded rather than lifted models for capabilities because agent capabilities exhibit non-liftable asymmetries (Sec.\,\ref{sec:empirical_results}). Formally,
\begin{definition}
\label{def:capability}
 A \emph{capability model} $h$ is a set of conditional-effect rules. Each rule $r \in h$ is a tuple $\langle \mathrm{cond}(r), \mathrm{effects}(r) \rangle$, where $\mathrm{cond}(r)$ is a Boolean formula over $V$ and $\mathrm{effects}(r) = \{(p_j, \eta_j)\}_j$
 is a probability distribution over effects. Each effect $\eta_j$ is a conjunction of literals over $V$, $p_j \in (0,1]$, and $\sum_j p_j = 1$.
\end{definition}

An effect $\eta=\langle\eta^+,\eta^-\rangle$ consists of sets of positive and negative literals, with $\mathrm{apply}(s,\eta)=(s\setminus\eta^-)\cup\eta^+$. The effect observed in transition $\langle s,c,s'\rangle$ is $\eta(\langle s,c,s'\rangle)=\langle s'\setminus s,s\setminus s'\rangle$. A transition $\langle s,c,s'\rangle$ is consistent with capability model $h$, denoted as $h\models\langle s,c,s'\rangle$, if there exists $r\in h$ such that $s\models\mathrm{cond}(r)$ and some $(p_j,\eta_j)\in\mathrm{effects}(r)$ satisfies $s'=\mathrm{apply}(s,\eta_j)$. Otherwise, $h\not\models\langle s,c,s'\rangle$.
A capability model $h_c$ for $c$ defines a conditional probability distribution over $c$'s outcomes: $\Pr_{h_c}(s'\mid s,c)=(1/|\mathcal{R}_s|)\sum_{r\in \mathcal{R}_s}\sum_{(p_j,\eta_j)\in\mathrm{effects}(r)}p_j\mathbf{1}[s'=\mathrm{apply}(s,\eta_j)]$ where $\mathcal{R}_s=\{r\in h_c:s\models\mathrm{cond}(r)\}$. If $|\mathcal{R}_s|=0$, then $\Pr_h(s'\mid s,c) = 0$. 
We denote $\hset_C$ as the set of capability models, one for each $c\in C$, and omit the subscript when clear from context.


    



A desired property of a learned capability model is that it is sound and complete with respect to a dataset of observed transitions $\mathcal{D}$. Formally,

\begin{definition}
Let $\alpha: X\rightarrow S$ be a representation function. A \emph{capability-transition dataset} $\mathcal{D}_{\mathcal{A},\mathcal{E}}$ is a multiset of triples $\langle s,c,s'\rangle$ where executing capability $c$ using BBAI $\mathcal{A}$ from state $x \in X$ results in $x' \in X$, with $s = \alpha(x)$ and $s' = \alpha(x')$.
\end{definition}

\begin{definition}
\label{def:soundness_completeness}
A model $h_c$ is \emph{sound w.r.t.\ }$\mathcal{D}$ if $\forall s,s' \in S,\ h_c \models \langle s,c,s'\rangle \Rightarrow \langle s,c,s'\rangle \in \mathcal{D}$, and \emph{complete w.r.t.\ }$\mathcal{D}$ if       $~\forall s,s' \in S,\ \langle s,c,s'\rangle \in \mathcal{D} \Rightarrow h_c \models \langle s,c,s'\rangle$.
\end{definition}

We omit ``w.r.t.\ $\mathcal{D}$'' when clear from context. Let $\mathcal{D}^\star$ denote all transitions in $S \times C \times S$ induced by the unknown true model $\hsettrue$. Requiring soundness w.r.t.\ $\mathcal{D} \subset \mathcal{D}^\star$ is overly restrictive, since models that generalize beyond observed transitions may violate it. Instead, we seek models that are complete w.r.t.\ $\mathcal{D}$ and sound w.r.t.\ $\mathcal{D}^\star$.
We use the variational distance (VD) to evaluate a hypothesis model $\hset$ w.r.t.\ $h^{\dagger\star}$ \citep{pasula2004learning}: $\emph{VD}(\hset, h^{\dagger\star}) =
(\sum_{(s,c,s') \in \mathcal{D}^\star}
|\Pr_{\hset}(s' \mid s,c) - \Pr_{h^{\dagger\star}}(s' \mid s,c)|)/|\mathcal{D}^\star|$.


Given a BBAI $\mathcal{A}$, the capability learning problem is to use a representation function and a simulator to discover $\mathcal{A}$'s capabilities and learn associated models that minimize VD from  $\mathcal{A}$'s (unknown) true capability model.

\section{Monte Carlo Query Synthesis }
\label{sec:mcqs}

In this section, we introduce \APPROACH\ (\APPABBREV), a query-synthesis algorithm for discovering a BBAI's capabilities and learning capability models. The complete algorithm, including the algorithm block, is provided in Appendix~\ref{sec:mcqs_alg}.
\APPABBREV\ reduces model uncertainty through falsification: it maintains a set of hypothesis models consistent with $\mathcal{D}$, synthesizes policies whose predicted outcomes disagree across those hypotheses, passes the policy as a \emph{query} to the BBAI, which then executes the policy and in the process creates a new trajectory that is added to
$\mathcal{D}$. Our approach works for capability definitions provided as a list of intents that the agent may be able to achieve. Alternatively, we can start with an empty set of capabilities and discover candidate capabilities before learning models for them.



\label{sec:capability-discovery}

We discover BBAI capabilities by letting it perform random walks in the simulator $\mathcal{S}_{\mathcal{E}}$ and collecting the resulting effects.  We further expand this set by observing state-capability transitions added to $\mathcal{D}$.
For each observed effect $\eta$, we extract every grounded atom $p(o_1,\ldots,o_k)$ that is added or removed and introduce the corresponding addition or deletion as an intent defining a new capability. If any grounding of a predicate $p$ appears in $\eta$, we assume that other type-consistent groundings of $p$ are likely achievable as well. Accordingly, \APPABBREV\ constructs capabilities for all intents $p(o'_1,\ldots,o'_k)$ where  $(o'_1,\ldots,o'_k)$ has the same type signature as the observed atom $p(o_1,\ldots,o_k)$. 

\subsection{Hypothesis Space Lattice for Query Synthesis}
\label{sec:hypothesis-space}

The central challenge in actively learning capability models is maximizing the information gained from each query. 
We show that the hypothesis space forms a lattice and, when restricted to hypotheses consistent with the state-capability transition dataset $\mathcal{D}$, admits two extremal hypotheses that bound all consistent hypotheses.

For state space $S$ and capability set $C$, let $T = S \times C \times S$ denote the set of possible transitions. Let compound hypothesis $h^\dagger$ denote a compiled set of hypotheses, one per capability in $C$, and let $\mathcal{H}$ denote the set of all compound hypotheses. $h^\dagger$ represents the subset of $T$ consistent with it: $T_{h^\dagger}=\set{t\in T:h^\dagger\models t}$. Two hypotheses $h^\dagger_1$ and $h^\dagger_2$ are equivalent, $h^\dagger_1\equiv h^\dagger_2$, iff $T_{h^\dagger_1}=T_{h^\dagger_2}$. This extension of $\models$ to compound hypotheses is well-defined because $h^\dagger$ contains exactly one hypothesis per capability. Consequently, the hypothesis space is partially ordered by inclusion: $h^\dagger_1\preceq h^\dagger_2\iff T_{h^\dagger_1}\subseteq T_{h^\dagger_2}$. This induces a subset lattice over $\mathcal{H}$.


Instead of considering the intractable and combinatorial hypothesis space exhaustively, we restrict our attention to hypotheses that are complete with respect to $\mathcal{D}$ (Def.\,\ref{def:soundness_completeness}). This restricted space has a unique minimal element: the lattice-meet of all complete hypotheses, which classifies only the transitions in $\mathcal{D}$ as consistent. This model is related to prior work on ``safe models'' \citep{ijcai2017p615}; we call it the \emph{pessimistic model} $h^\dagger_\land$ to emphasize that it permits only directly observed transitions.

As noted earlier, soundness w.r.t.\ $\mathcal{D}$ is too restrictive for filtering hypotheses. We therefore relax it to soundness modulo unseen capability executions: a model $h^\dagger$ is \emph{possibly sound} w.r.t.\ $\mathcal{D}$ iff, whenever $h^\dagger \models \tuple{s,c,s'}$, either $\tuple{s,c,s'}\in\mathcal{D}$, or the following conditions hold: (a) $\forall s'',  \tuple{s,c,s''}\not \in\mathcal{D}$, and (b) $\exists s_1,s_2, \tuple{s_1,c,s_2}\in\mathcal{D}$ s.t. $\eta(s,s')=\eta(s_1,s_2)$. This notion of soundness allows for the possibility that, once an execution of $c$ in $s$ is observed, $h^\dagger$'s prediction will prove sound. We define the \emph{optimistic model} $h^\dagger_\lor$ as the join of all possibly sound models w.r.t.\ $\mathcal{D}$.

Two properties of $\mathcal{D}$ ensure that pessimistic and optimistic models bound the true model.  $\mathcal{D}$ is \emph{transition complete} iff for all $s,c$ whenever there exists $s'$ such that $\langle s,c,s'\rangle\in\mathcal{D}$, then for all $s''$ s.t. $h^{\dagger\star}\models\langle s,c,s''\rangle$, $\langle s,c,s''\rangle\in\mathcal{D}$. We say that $\mathcal{D}$ is \emph{effect complete} iff for all $s,c$, whenever $h^{\dagger\star}\models\langle s,c,s'\rangle$, then there exists $\langle s_1,c,s_2\rangle\in\mathcal{D}$ such that $\eta(s,s')=\eta(s_1,s_2)$. 
These conditions ensure the bounds (Appendix~\ref{sec:formal_proofs} includes proofs for all theoretical results):

\begin{restatable}{lemma}{envlemma}
\label{lemma:env_lemma}
Let $C$ be a set of agent capabilities with true model $h^{\dagger\star}$ expressible over vocabulary $V$, and let $h^\dagger_\land$ and $h^\dagger_\lor$ be the pessimistic and optimistic models constructed
from transition data $\mathcal{D}$. If $\mathcal{D}$ is transition and effect complete, then $\hset_\land \preceq \hsettrue \preceq \hset_\lor$.
\end{restatable}

Furthermore,  $h^\dagger_{\land}$ and $h^\dagger_{\lor}$ can be computed efficiently from $\mathcal{D}$ as follows. For each capability $c$, let $\mathcal{D}_c=\set{ \langle s,c,s'\rangle:\exists s, s'~\tuple{s,c,s'}\in\mathcal{D}}$ denote the observed transitions involving $c$, and let 
$\eta_c(s,\mathcal{D})=\set{\eta(\langle s,c,s'\rangle):\exists s'~ \langle s,c,s'\rangle\in\mathcal{D}_c}$
denote the effects observed from state $s$ under $c$. We partition states by equality of these effect sets: $s_1\sim_c s_2$ iff $\eta_c(s_1,\mathcal{D})=\eta_c(s_2,\mathcal{D})$, yielding partitions $\Phi_c$. Each block $S_\varphi\in\Phi_c$ has a common  effect set $E_\varphi$ and induces one conditional effect rule.
%
%
Let $\ell(s)=\bigwedge_{p\in s}p\wedge\bigwedge_{p\notin s}\neg p$ denote representation of a state as a formula. 

We then derive the optimistic and pessimistic models using BDDs by extending effect partitioning~\citep{mordoch2024safe} to stochastic effects. A set $S_\varphi$ where a certain effect occurs can be captured with a pessimistic condition $\mathrm{pcond}(S_\varphi)=\bigvee_{s\in S_\varphi}\ell(s)$ or an optimistic condition
$\mathrm{ocond}(S_\varphi)=\neg[\bigvee_{s\in \Phi_c\setminus{S_\varphi}}\ell(s)]$. 
The corresponding rule is $r_\varphi=\langle \mathrm{cond}(S_\varphi),{(\Pr(\eta),\eta):\eta\in E_\varphi}\rangle$, where $\mathrm{cond}$ is either $\mathrm{pcond}$ or $\mathrm{ocond}$ represented as a BDD and $\Pr(\eta)$ is estimated from $\mathcal{D}_c$ by MLE.  Collecting all rules for all capabilities using the  pessimistic conditions yields $\hset_{\land}$, while using optimistic conditions yields $\hset_{\lor}$. This gives two hypotheses bounding the space of consistent hypotheses,  enabling \APPABBREV\ to search for queries maximizing their disagreement.

\subsection{MCQS Using Lattice-Based Learning}
\label{sec:mcts_query_synthesis}

We now present the overall process for query synthesis based on the lattice theory developed above.
For  hypotheses $\hset_{1}$ and $\hset_{2}$ over the capability set $C$, a capability policy $\pi:S\rightarrow C$ is \emph{informative} from an initial state $x_0\in X$ if executing $\pi$ from $x_0$ 
results in disparate outcome distributions under $\hset_1$ and $\hset_2$.
Once such a policy is identified, we send it as a query to the agent.
Since the environment may be stochastic, a query requires  multiple independent executions of $\pi$ from the same initial state, 
thereby increasing the probability of observing an informative outcome. Formally:

\begin{definition}
\label{def:query}
    A \emph{query} is a tuple $\langle x_0,\pi,n\rangle$, where $x_0\in X$ is an initial environment state, $\pi:S\rightarrow C$ is a policy mapping  represented states to capabilities, and $n\in\mathbb{Z}^+$ is the number of executions of $\pi$.
\end{definition}

The query-synthesis objective is to identify a policy
whose execution induces different predicted state distributions under two capability-set hypotheses. For a capability $c_i$, let $\tau_i^{\mM}(\rho)$ denote the state distribution predicted by model $\mM$ after executing $c_i$ from an initial distribution $\rho$: $\tau_i^{\mM}(\rho)(s') = \sum_{s\in S} \Pr_{\mM}(s'\mid s,c_i)\rho(s)$. We write $\tau_{1,\ldots,k}^{\mM}(\rho)$ for the distribution induced by executing the capability sequence $c_1,\ldots,c_k$ from $\rho$. 
Let $\hat{\delta}$ be a distance function over distributions and $\delta$ a threshold. An informative  policy $\pi$ is one under which  $\hat{\delta}(\tau_{\pi}^{\hset_{1}}(\rho_0), \tau_{\pi}^{\hset_{2}}(\rho_0)) > \delta$.

We formulate this objective as a Markov decision process over pairs of predicted state distributions. Let $\rho_1$ and $\rho_2$ denote the state distributions maintained under $\hset_{1}$ and $\hset_{2}$, respectively. States for the query synthesis MDP are meta-states of the form $\langle\rho_1,\rho_2\rangle$; its actions select a capability $c\in C$; applying $c$ on a meta-state updates each distribution according to its corresponding model. Formally:

\begin{definition}\label{def:dist_mdp}
 The \emph{distinguishing MDP} $\mathcal{P}^\dagger(s_0,\hset_{1},\hset_{2})$, where $s_0\in S$ is the  initial state and $\hset_{1}$ and $\hset_{2}$ are capability models, is a tuple $\langle \mathcal{S}^\dagger, A^\dagger, \mathcal{T}^\dagger, R^\dagger, s_0^\dagger \rangle$, where $\mathcal{S}^\dagger = \{\, \langle \rho_1, \rho_2 \rangle \mid \rho_1,\rho_2 \text{ are probability distributions over } S\,\}$; $s_0^\dagger = \langle \rho_{1,0}, \rho_{2,0}\rangle$ with $\rho_{1,0}(s_0)=1$ and $\rho_{2,0}(s_0)=1$; $A^\dagger = C$ is the set of capabilities; $\mathcal{T}^\dagger(\langle \rho_1, \rho_2\rangle, c_i)=\langle \tau^{\hset_{1}}_i(\rho_1), \tau^{\hset_{2}}_i(\rho_2)\rangle$; and $R^\dagger(\langle \rho_1, \rho_2\rangle)= \hat{\delta}(\rho_1, \rho_2)$.
\end{definition}

\begin{figure*}[t]
    \centering
    \includegraphics[width=\textwidth]{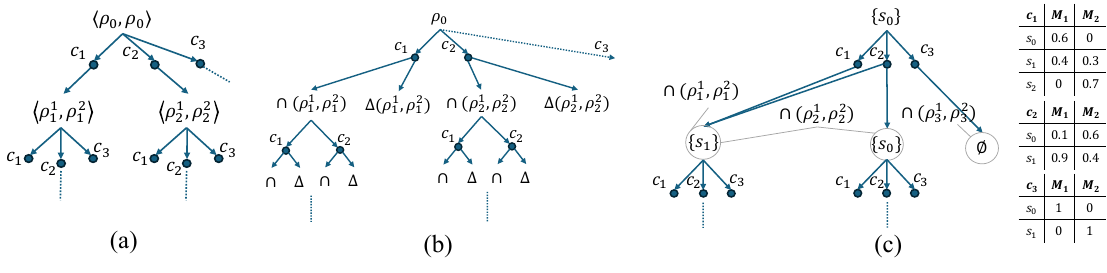}
    \caption{\small Overview of MCQS. $\rho^i_j$ denotes $\tau^{M_i}_j(\rho_0)$; $\cap()$ and $\Delta()$ represent the intersection and the symmetric difference of sets of support (SoS) of their input distributions, respectively. (a) MCTS formulation of the Distinguishing MDP (Def.\ref{def:dist_mdp}); (b) an SoS based representation that replaces each pair of outcome distributions into the intersection and symmetric difference of their supports, and does not generate children for the symmetric difference as they are already distinguished.
    (c) a sample-based approximation of the SoS representation, where each node represents a sample from two distributions' support sets. Tables on the right show  outcome probabilities under the two models.}
    \label{fig:mcts}
\end{figure*}

Solving a distinguishing MDP amounts to finding a policy $\pi$ that maximizes predicted disagreement: $\arg\max_\pi \hat{\delta}(\tau_{\pi}^{\hset_{1}}(\rho_0), \tau_{\pi}^{\hset_{2}}(\rho_0))$. To find $\pi$, we use MCTS with UCT~\citep{kocsis06_mcts,swiechowski2023monte}. Let $Q(s^\dagger,c)$ denote the value for applying capability $c$ in state $s^\dagger$, and let $N(s^\dagger)$ and $N(s^\dagger,c)$ denote node and edge visit counts. With exploration constant $\lambda_{\mathrm{uct}}$, we define $\mathrm{UCT}(s^\dagger,c)=Q(s^\dagger,c)+\lambda_{\mathrm{uct}}\sqrt{\ln N(s^\dagger)/N(s^\dagger,c)}$. During selection, expansion, and policy extraction, MCTS selects $\arg\max_{c\in C}\mathrm{UCT}(s^\dagger,c)$. Retaining the exploration bonus in the query encourages policies targeting rarely explored branches. We develop two MCQS variants that differ in state-distribution representations and rewards.

\label{sec:prob_circuits}
\emph{MCQS-Exact} (MCQS-E, Fig.~\ref{fig:mcts}(a)) solves the distinguishing MDP directly by maintaining explicit predicted distributions over represented states, encoding each state as a bit-vector with one bit per literal and annotating it with its probability mass. Bit masks test whether a state satisfies each condition in the conditional-effects rules and apply the corresponding effects, yielding an exact but compact representation of the distribution pairs propagated by MCTS (Appendix\,\ref{sec:exact_implementation_details}). MCQS-E uses total variation distance $\delta_{\mathrm{TV}}$ for $\hat{\delta}$ since it is symmetric, bounded, and rewards policies assigning probability mass to states likely under one model but unlikely under the other. For node $n$ representing distinguishing-MDP state $s^\dagger$, backpropagation uses the best expected disagreement along the path: $Q(n,c)=\max(\delta_{\mathrm{TV}}(s^\dagger),V(\mathcal{T}^\dagger(s^\dagger,c)))$ and $V(n)=\max_{c\in C}Q(n,c)$.

\emph{MCQS with set-of-support (SoS) factorization} uses the observation that 
a state in the symmetric difference of the two models' support sets, denoted by $\Delta(\cdot)$ in Fig.\,\ref{fig:mcts}(b), is deterministically falsifying: observing such a state in the execution rules out the model whose support excludes it. Such states therefore need not be propagated through MCTS; only states in the \emph{intersection} of the two supports, denoted by $\cap(\cdot)$, carry residual ambiguity. This observation motivates \APPABBREV-S, which propagates only intersection states and approximates them via sampling.

\label{sec:cat_mcts}

\emph{\APPABBREV-S} is sample-based MCQS (Fig.~\ref{fig:mcts}(c)). It builds on the SoS formulation by replacing nodes that represent intersections of support sets with state nodes sampled from those intersections.
At each node with state $s\in S$, \APPABBREV-S selects a valid capability from $C_s=\{c\in C\mid \exists r\in \mathrm{cond\_effects}(c),\, s\models \mathrm{cond}(r)\}$. It then simulates the result of executing capability $c$ by sampling a successor state from the intersection of the two predicted support sets: $s'\sim 0.5\Pr_{h_1}(s'\mid s,c)+0.5\Pr_{h_2}(s'\mid s,c)$. We use  $\Delta(\rho_1, \rho_2)$ 
to refer to the symmetric difference of the support sets of $\rho_1$ and $\rho_2$. Following intuition behind SoS factorization, we define an approximation of TV distance that focuses on states in the symmetric difference: $\hat{\delta}(\rho_1, \rho_2)=\delta_{SD}(\rho_1,\rho_2)=E[I_{\Delta(\rho_1, \rho_2)}]=\sum_{s\in \Delta(\rho_1, \rho_2)}0.5\rho_1(s)+0.5\rho_2(s)$. This is used in sample based $Q$ estimates as follows. 
 
During tree traversal, $Q$ is computed as $Q(s,c)=R(s)+\sum_{s'}\Pr(s'\mid s,c)V(s')$, where $V(s)=\max_{c\in C_s}Q(s,c)$. Let $c_i$ be the capability leading to node $n$ with state $s$, and let $\langle\rho_1,\rho_2\rangle$ be the distribution pair at $n$'s parent. Then $R(s)=1$ if $s\in\Delta(\tau_i^{h_{C,1}}(\rho_1),\tau_i^{h_{C,2}}(\rho_2))$, and $R(s)=0$ otherwise. When a state-capability sequence $s,c,s',c'$ is observed during one traversal (including the rollout from a leaf), $Q$ is updated using just the observed state $s'$ rather than the expectation over all states.

\subsection{Robust Query Synthesis and Execution}
\label{sec:optimizations}
The distinguishing-MDP formulation specifies an ideal query-synthesis objective. We now describe the optimizations \APPABBREV uses to execute queries robustly, revisit informative state-capability pairs, and update the observation dataset. 
Both \APPABBREV-E and \APPABBREV-S  allow query execution to adapt to observed outcomes (Appendix~\ref{sec:dynamic_policy}). Additionally, to encourage exploration, we sample the initial state for query synthesis and execution based on frequency (Appendix~\ref{sec:init_state_sample}).

To encourage state-capability revisitation, \APPABBREV uses a $\xi$-exploration bonus that assigns probability mass to potentially missing effects. Let $\mathcal{D}_{SC}(s,c)$ denote the observations collected for state-capability pair $(s,c)$. We estimate the probability of a missing effect as $\Pr_m(s,c)=1/(1+|\mathcal{D}_{SC}(s,c)|)$. When evaluating a query, \APPABBREV assigns probability $1-\xi\Pr_m(s,c)$ to the observed effect distribution and probability $\xi\Pr_m(s,c)$ to unseen distinguishing effects. Thus, larger values of $\xi$ encourage revisiting previously tried state-capability pairs, while smaller values prioritize discovering new state-capability pairs.


\label{sec:data}

The dataset $\mathcal{D}$ is updated as follows. We define the \emph{temporal length} of an execution trajectory as the number of symbolic state changes and  limit agent execution per capability using a bound $\Theta$ on this temporal length. When executing a capability $c$ results in a trajectory  $\overline{x}=\langle x_0,x_1,\ldots,x_k\rangle$ that abstracts into a represented state trajectory $s_0, s_1, \ldots, s_j$ ($j< \Theta$) where all consecutive identical states are removed, $\tuple{s_0, c,  s_j}$ is added to $\mathcal{D}$.


\section{Results}
\label{sec:results}


\subsection{Theoretical results}
\label{sec:theory}

We prove that under realizability and sufficient exploration, \APPABBREV{} converges to the true capability model in the limit of infinite samples; proofs and results appear in Appendix\,\ref{sec:formal_proofs}. If MCTS returns optimal policies, we can prove the following.

\begin{restatable}{theorem}{mainthm}
\label{theorem:vd_limit}
Let $C$ be a set of agent capabilities whose true model $h^{\dagger\star}$ is expressible over vocabulary $V$, and let $\hset_i$ denote the model learned by \APPABBREV{} after iteration $i$. If $\xi>0$, then $\mathrm{VD}(\hset_i,h^{\dagger\star})\xrightarrow[i\to\infty]{\mathrm{a.s.}} 0$
over all reachable transitions induced by capabilities in $C$.
\end{restatable}




We additionally prove that there exists a finite set of transitions whose observation is sufficient for the pessimistic and optimistic models to become equivalent.

\begin{restatable}{theorem}{supportthm}
\label{theorem:finite_transition_diff}
Let $\mathcal{D}$ be the transitions observed by \APPABBREV{}. If the true agent model $h^{\dagger\star}$ is expressible over vocabulary $V$, then there exists a finite set of transitions $T_\delta$ such that the pessimistic and optimistic models computed from $\mathcal{D} \uplus T_\delta$, denoted by $h^{\dagger}_\land$ and $h^{\dagger}_\lor$, respectively, satisfy $h^{\dagger}_\land \equiv h^{\dagger}_\lor$.
\end{restatable}

We prove that $h^\dagger_\land$ is sound and complete with respect to its dataset, while $h^\dagger_\lor$ is complete (Theorem\,\ref{theorem:opt_pess_sc}). For a transition and effect complete dataset, $h^\dagger_\land\equiv h^\dagger_\lor$ guarantees equivalence to the true model (Theorem\,\ref{theorem:termination}) and structural equivalence of any complete and possibly sound model to the true model (Theorem\,\ref{theorem:all_sc_model}). Proofs in Appendix\,\ref{sec:formal_proofs}.

\subsection{Empirical Evaluation}
\label{sec:empirical_results}
Our empirical evaluation addresses four questions: \emph{(E1) Does \APPABBREV{} learn accurate capability models more efficiently than other approaches?} \emph{(E2) Can \APPABBREV{} reveal useful information about the limits and capabilities of SOTA BBAIs?} \emph{(E3) Do the learned capability models accurately predict reachability?} and \emph{(E4) Is maximizing disagreement alone sufficient for learning accurate capability models?} Hyperparameter configurations and detailed results are presented in Appendices\,\ref{sec:emp_add_info} and \ref{appendix:add_results}.

We assess \S E1 by comparing models learned with \APPABBREV{}-S, \APPABBREV{}-E, the closest existing approach, and a random-querying ablation baseline across several agents and environments. For \S E2, we inspect \APPABBREV{}'s learned models for notable BBAI capabilities and limitations. For \S E3, we evaluate whether models learned by \APPABBREV{} accurately predict outcomes of capability sequences. For \S E4, we ablate the revisitation reward by setting $\xi=0$.

\paragraph{Agents}
We evaluate six agents across several environments. The HDDLGym~\citep{La_Mon-Williams_Shah_2025} RL agent operates in the Overcooked domain~\citep{carroll2019utility}. The ReAct agent uses GPT-5.4-nano~\citep{gpt54nano} in a MiniGrid environment~\citep{gym_minigrid}. Both CSteve~\citep{park2025crafterdojo} and Qwen CSteve operate in a Crafter environment~\citep{hafner2022benchmarking}. Finally, the LAO* agent operates on PDDLGym domains~\citep{silver2020pddlgym}, including Blocksworld and First Responders. As an application, we also evaluate the LLM-based robotic agent SayCan~\citep{ahn2022saycan}.

\paragraph{Baselines}
To our knowledge, no prior method performs active capability discovery and learning for BBAIs, so we compare against the closest existing approach, QACE~\citep{verma23neurips}, for \S E1, and two \APPABBREV{} ablations for \S E1 and \S E4. QACE requires the agent's capabilities and intents as input; we supplied those learned by \APPABBREV{}. QACE learned an accurate model for LAO* on Blocksworld, but produced no model for the other problems within 50 hours due to the limited expressiveness of its model class. The \emph{Random Query} ablation replaces MCTS query synthesis with a uniformly sampled capability sequence $\pi=(c_1,c_2,\ldots,c_{30})$, where $c_i\sim\mathrm{Uniform}(C)$, leaving all other components unchanged and isolating the contribution of query synthesis. For \S E4, we ablate \APPABBREV{} by setting $\xi=0$.

\begin{figure*}
    \centering
    \includegraphics[width=1\linewidth]{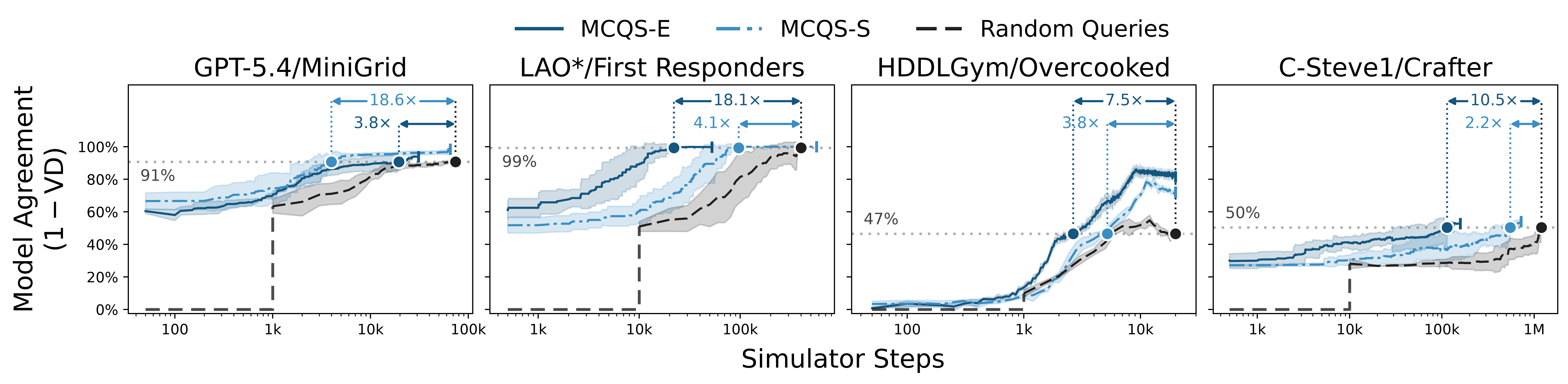}
    \caption{Model agreement (1 - weighted VD) of \APPABBREV{}-E and \APPABBREV{}-S on four evaluation problems. Shaded regions indicate one standard deviation over multiple runs. Annotated arrows indicate the simulator step-count ratio between each \APPABBREV{} approach and the Random Query ablation baseline.}
    \label{fig:vd_results}
\end{figure*}

\paragraph{Evaluation metrics and methodology}
We evaluate how accurately a learned model $\hset$, constructed from $\mathcal{D}_{h^\dagger}$, predicts observed transition dynamics using weighted sampled variational distance $\mathrm{VD}{s}(\hset,h^{\dagger\star})$. To construct an independent evaluation dataset $\mathcal{D}'$, we execute the BBAI $\mathcal{A}$ using the same environment and representation function, collecting 100,000 sequences of 10--30 capabilities (5,000 for MiniGrid). Let $N_e(s,c,s'\mid\mathcal{D})$ denote the number of occurrences of transition $\langle s,c,s'\rangle$ in $\mathcal{D}$ and $N_e(s,c\mid\mathcal{D})=\sum_{s'}N_e(s,c,s'\mid\mathcal{D})$. Let $T_{\mathcal{D}'}$ denote the unique transitions observed in $\mathcal{D}'$, and $c_T=\sum_{\langle s,c,s'\rangle\in T_{\mathcal{D}}}N_e(s,c,s'\mid\mathcal{D}_{h^\dagger})$ be the number of transitions observed in $\mathcal{D}_{h^\dagger}$. We define $\mathrm{VD}_{s}(\hset,h^{\dagger\star})$ as
$
   \mathrm{VD}_{s}(\hset,h^{\dagger\star})
=\frac{1}{c_T}\sum_{\langle s,c,s'\rangle\in T_{\mathcal{D}'}}
N_{e}(s,c,s'\mid\mathcal{D}_{h^\dagger})\times\left|
\frac{N_{e}(s,c,s'\mid\mathcal{D}')}{N_{e}(s,c\mid\mathcal{D}')}
-\Pr_{\hset}(s'\mid s,c)
\right|.
$

We evaluate learned pessimistic models because they are safer for unseen states. To focus on achievable intents, we remove conditional-effect rules that fail to achieve their intent and discard capabilities never observed as achievable. We use temporal length $\infty$ for MiniGrid, SayCan, and Overcooked, and $1$ otherwise.

We perform five independent model-learning runs per problem. \APPABBREV{} stops after a timeout or 25 consecutive queries without a newly discovered transition or significant model change (an effect probability changing by more than $5\%$). To balance revisitation with hard-to-reach capability discovery, we dynamically set $\xi$ (Sec.\,\ref{sec:optimizations}) based on the number of queries since the last discovery, up to $10^{-5}$. Timeouts are at most two days. 

\subsubsection{Results and Analysis} We now present our analysis of results addressing the evaluation questions E1--E4 listed above.

\noindent\emph{\S E1 (Learning efficiency)} Fig.~\ref{fig:vd_results} compares model agreement (1 - weighted VD) over the number of simulator steps. On MiniGrid, \APPABBREV{}-S reaches $91\%$ model agreement with $18.6\times$ fewer simulator steps than the random-query baseline. On First Responders, \APPABBREV{}-S reaches over $99\%$ with $18.1\times$ fewer steps. On Overcooked, in 10,000 simulator steps, \APPABBREV{}-E achieves nearly double the model agreement of the random-query baseline.

Crafter's coarse representation function aggregates grounded states sharing the same symbolic state, increasing VD and the simulator steps required to reduce it. This trend is consistent across both CSteve variants, so we report only one. Despite this challenge, \APPABBREV{}-E reaches 50\% model agreement with 10.5$\times$ fewer simulator steps than random querying, suggesting that \APPABBREV{} remains effective under imperfect representation functions.




\noindent\emph{\S E2 (Limits and capabilities of SOTA BBAIs)} Fig.~\ref{tab:capabilities} highlights unintuitive limitations discovered by \APPABBREV{}-E that are readily interpretable and can inform BBAI development. GPT-5.4 reliably picks up a key when requested, but often unnecessarily traverses to and opens a door; moreover, asking it to pick up the key and then unlock the door is less successful than directly requesting it to unlock the door. SayCan is highly stochastic, with many desired effects occurring with low probability; for example, it stacks a green block on a yellow block only 6\% of the time, often misidentifying the yellow block as green. Finally, LAO*'s place-block-A capability reveals a preference likely induced by internal tie-breaking and stationary policies.

These results also illustrate the challenges of lifting accurate capability models for BBAIs. We considered lifted models, but emergent agent capabilities often lack symmetries under object substitution. For example, the LAO* agent prefers placing blocks on C rather than B to clear its hand, while the MiniGrid agent can retrieve a key from NW or SW, but not from other quadrants. Lifting loses such object-specific dependencies, producing exaggerated capabilities that misrepresent the agent's true behavior.

\noindent\emph{\S E3 (Cross evaluation through reachability)} We evaluated \APPABBREV{}-S models on 100 predicted-reachable 5-capability sequences, executing each 100 times on the BBAI. Models achieved $81\%\pm3.6\%$ accuracy on MiniGrid, $100\%$ on First Responders, and $78\%\pm16\%$ on Crafter. These results demonstrate that learned capability models can predict downstream performance and support planning. 

\noindent\emph{\S E4 (Ablation w.r.t. visitation parameter)} With $\xi=0$, the resulting models produced by \APPABBREV{} achieved higher than 98\% 
model agreement
across three PDDLGym problems: Blocksworld, Tireworld, and First Responders. This demonstrates the effectiveness of rewarding disagreement alone. The revisitation incentive had mixed effects on accuracy and efficiency, with \APPABBREV{}-S being more sensitive to $\xi$. 


\section{Related Work}
\label{sec:related_work}
Research on action model learning focuses on dynamic models of the environment rather than of the agent~\citep{pasula2007learning,juba_22_learning,gosgens2025learning,verma2021asking,benyamin2025integrating}. This line of work does not address the problem of learning capability models that capture an agent's decision making capabilities, which is the focus of this paper. 
\citet{tantakoun2025llms} present a survey of  LLM-to-PDDL approaches for world model learning, and recent benchmarks~\citep{hu2025text2world,zuo-etal-2025-planetarium} evaluate model generation from LLMs. 
Recent work also addresses the problem of learning capability models~\citep{verma2022discovering,shah2025from} in the limited setting of deterministic models with  add and delete effects.  
In contrast, our approach learns rich probabilistic capability models conditioned on arbitrary logical formulas in stochastic settings.
Testing and verification approaches such as DeepXplore~\citep{pei2017deepxplore}, Metamorphic Testing~\citep{chen2018metamorphic}, safety verification~\citep{tran2019safety,dreossi2019verifai,araujo2023testing}, etc.\ aim to find failures or verify properties, rather than discover what agents can do. Behavior modeling and inverse planning~\citep{baker2009action,shvo2020activegoal} focus on inferring agent goals and strategies from observations, whereas specification and invariant learning~\citep{leucker2009runtime,neider2018invariant,bao2024data} extract system constraints. 
While these areas share the goal of understanding black-box systems, our approach differs by discovering and modeling the capabilities an agent can reliably execute under different conditions.

\section{Conclusions and Future Work}
We presented \APPABBREV{}, an active query-synthesis approach for discovering and modeling BBAI capabilities in stochastic settings. Empirical and theoretical results show that \APPABBREV{} efficiently learns useful, accurate capability models within a finite query budget. While effective, \APPABBREV{} leaves open improvements to exploration efficiency and generalization. In particular, real-world BBAIs often exhibit implicit or context-dependent preferences among multiple valid plans. Distinguishing such preferences from structural constraints remains an open challenge for learning models that both generalize and accurately reflect an agent's behavior.

\section*{Acknowledgments}
This work was supported in part by the following grants:
NSF IIS 2419809, ONR N00014-23-1-2416, and AFOSR
FA9550-25-1-0320. We thank Ruthvick Suresh for contributions to the empirical evaluation in an earlier version of this work. We also thank Sai Saketh Reddy Kambalapally for contributing the Crafter agents evaluated in this paper.

\bibliography{references}

\appendix


\section{Formal Proofs}
\label{sec:formal_proofs}

This appendix provides proofs of Lemma\,\ref{lemma:env_lemma} (Sec.\,\ref{sec:hypothesis-space}) and theorems presented in Sec.\ref{sec:theory}, along with additional theoretical results. As discussed in Sec.\,\ref{sec:theory}, the proof of Theorem~\ref{theorem:vd_limit} requires the assumption \emph{A1: MCTS returns an optimal policy}. This assumption is used only in the proofs of Lemmas~\ref{lemma:revisiting} and~\ref{lemma:revisit_improvement}, which in turn are used exclusively to establish Theorem~\ref{theorem:vd_limit}. All remaining lemmas and theorems are proved without relying on Assumption~A1.

We begin by proving that, with respect to the dataset from which they are constructed, the pessimistic model is sound and complete, while the optimistic model is complete.

\begin{theorem}
\label{theorem:opt_pess_sc}
Let $\mathcal{D}$ be the observed transitions, and let $h^\dagger_\land$ and
$h^\dagger_\lor$ be the pessimistic and optimistic models computed by
\APPABBREV, respectively. Then $h^\dagger_\land$ is sound and complete with respect to
$\mathcal{D}$, and $h^\dagger_\lor$ is complete with respect to $\mathcal{D}$.
\end{theorem}

\begin{proof}
Let $\langle s,c,s'\rangle \in \mathcal{D}$. By construction
$h^\dagger_\land\models\langle s,c,s'\rangle$, so $h^\dagger_\land$ is complete w.r.t.\ $\mathcal{D}$ (Def.\,\ref{def:soundness_completeness}). Since every transition admitted by $h^\dagger_\land$ lies in $\mathcal{D}$, it is possibly sound, and as $h^\dagger_\lor$ is the join of all possibly sound hypotheses, $h^\dagger_\land\preceq h^\dagger_\lor$. Hence $h^\dagger_\lor\models\langle s,c,s'\rangle$ and $h^\dagger_\lor$ is complete w.r.t.\ $\mathcal{D}$ (Def.\,\ref{def:soundness_completeness}).

Conversely, if $h^\dagger_\land\models\langle s,c,s'\rangle$ then $\langle s,c,s'\rangle\in\mathcal{D}$ by construction, so $h^\dagger_\land$ is sound w.r.t.\ $\mathcal{D}$ (Def.\,\ref{def:soundness_completeness}).
\end{proof}

\subsection{Bounding the True Model}
\label{sec:bound_model_proof}

Let $T_{h^{\dagger\star}}$ denote the set of transitions consistent with the true model. We first show that $h^{\dagger\star}$ is bounded by the pessimistic and optimistic hypotheses when $h^{\dagger\star}$ is expressible with the vocabulary and $\mathcal{D}$ is transition and effect complete (Sec.\,\ref{sec:hypothesis-space}).

\envlemma*
\begin{proof}
By construction, $h^\dagger_\land$ admits exactly the observed transitions $T_\mathcal{D}$, so $T_{h^\dagger_\land}=T_\mathcal{D}$. Since observed transitions are consistent with the
true model, $T_\mathcal{D}\subseteq T_{h^{\dagger\star}}$, giving $\hset_\land \preceq \hsettrue$.

For the upper bound, let $\langle s,c,s'\rangle\in T_{\hsettrue}$. If $\langle s,c,s'\rangle\in T_\mathcal{D}$, then $\langle s,c,s'\rangle\in  T_{h^\dagger_\lor}$ by Theorem~\ref{theorem:opt_pess_sc}.
Otherwise $(s,c)$ is untried: were it tried, the premise of $\mathcal{D}$ being transition complete means that effect $h^{\dagger\star}$ produces at $(s,c)$ was observed, i.e. $\langle s,c,s'\rangle\in T_\mathcal{D}$. Let $\eta$ be the effect realizing $s'$ at $s$, so $\mathrm{apply}(s,\eta)=s'$. By the premise of $\mathcal{D}$ being effect complete, $\eta$ has been witnessed in $\mathcal{D}$, so $h^\dagger_\lor$ contains a rule $r$ with $\eta\in\mathrm{effects}(r)$. By construction, the optimistic condition $\mathrm{ocond}$ excludes a state only if that state is observed to be governed by a different rule; since $(s,c)$ is untried, no observation assigns $s$ to a rule for $c$, so $s\models\mathrm{ocond}(r)$ and $\langle s,c,s'\rangle\in T_{h^\dagger_\lor}$.

Hence $T_{h^{\dagger\star}}\subseteq T_{h^\dagger_\lor}$, so
$\hsettrue \preceq \hset_\lor$ and $\hset_\land \preceq \hsettrue \preceq \hset_\lor$.
\end{proof}

\subsection{MCQS Convergence to the True Model}

We now prove the main theoretical result from Sec.\,\ref{sec:theory}: the variational distance between a model $h^\dagger$ produced by \APPABBREV{} and the true model $h^{\dagger\star}$ almost surely converges to zero on reachable transitions as the number of iterations of \APPABBREV{} approaches infinity.

We first characterize the reward when the distributions $\rho_0$ and $\rho_1$ represented by a meta-state $s^\dagger$ are equal. Specifically, letting $\rho_0=\rho_1=\rho$, if capability $c$ has previously been executed in every state in $\operatorname{supp}(\rho)$, we show that the reward of the resulting meta-state $s_1^\dagger$ is exactly the expected revisitation bonus over its constituent states.

\begin{lemma}
\label{lemma:reward}
Let $c\in C$ be a capability and $s^\dagger$ be a meta-state of the distinguishing MDP $\mathcal{P}(s_0,h^\dagger_\land,h^\dagger_\lor)$ representing distributions $\rho_0$ and $\rho_1$. If $\rho_0=\rho_1=\rho$ and, for every $s\in\operatorname{supp}(\rho)$, the transition dataset $\mathcal{D}$ contains an observation of executing $c$ from $s$, then the reward of the resulting meta-state of executing $c$ in $s^\dagger$, $s_1^\dagger$, is $R(s_1^\dagger)=\sum_{s\in {supp}(\rho)}\frac{\rho(s)\xi}{\mathcal{D}_{SC}(s,c)+1}$.
\end{lemma}

\begin{proof}
By the premises, for each state $s\in\operatorname{supp}(\rho)$, $\mathcal{D}$ contains an observation of performing $c$ in $s$. By construction, the pessimistic and optimistic models therefore predict the same transitions from performing $c$ in $s$ and thus induce the same successor distribution. Since $\rho_0=\rho_1=\rho$, the two distributions $\rho'_0$ and $\rho'_1$ represented by  $s_1^\dagger$ are equal, and hence $\hat{\delta}(\rho'_0,\rho'_1)=0$.

Thus, the only remaining reward for each state $s$ is its revisit value, $\frac{\xi}{\mathcal{D}_{SC}(s,c)+1}$. Weighting each revisit value by $\rho(s)$, the reward from performing $c$ in $s^\dagger$ is $R(s_1^\dagger)=\sum_{s\in {supp}(\rho)}\frac{\rho(s)\xi}{\mathcal{D}_{SC}(s,c)+1}$.
\end{proof}

We define consistency in terms of variational distance. Two models, $h_1^\dagger$ and $h_2^\dagger$, are \emph{consistent} w.r.t. state-capability pair $(s,c)$ iff $\forall s',\Pr_{h_1^\dagger}(s'|s,c)=\Pr_{h_2^\dagger}(s'|s,c)$. Therefore,  $h_1^\dagger$ and $h_2^\dagger$ are inconsistent w.r.t. $(s,c)$ iff $\exists s',\Pr_{h_1^\dagger}(s'|s,c)\neq\Pr_{h_2^\dagger}(s'|s,c)$. The inconsistency with $h_1^\dagger$ is reduced between models $h_2^\dagger$ and $h_3^\dagger$ w.r.t. $(s,c)$ iff $\sum_{s'\in S}|\Pr_{h_3^\dagger}(s'|s,c)-\Pr_{h_1^\dagger}(s'|s,c)| < \sum_{s'\in S}|\Pr_{h_2^\dagger}(s'|s,c) - \Pr_{h_1^\dagger}(s'|s,c)|$.

We next show that any reachable state-capability pair $(s,c)$ on which either $h^\dagger_\land$ or $h^\dagger_\lor$ is inconsistent with $h^{\dagger\star}$ is eventually queried with non-zero probability.

\begin{lemma}
\label{lemma:revisiting}
Let $h^{\dagger\star}$ be the true model, expressible over vocabulary $V$, and let $h^\dagger_\land$ and $h^\dagger_\lor$ be the pessimistic and optimistic models computed by MCQS, respectively. If $\xi>0$ and there exists a reachable state-capability pair $(s,c)$ such that either $h^\dagger_\land$ or $h^\dagger_\lor$ is inconsistent with $h^{\dagger\star}$ w.r.t. $(s,c)$, then, after finitely many iterations, MCQS produces a query with non-zero probability of executing $c$ in $s$.
\end{lemma}

\begin{proof}
Since meta-states exactly represent the induced distributions, any finite capability sequence deterministically induces a sequence of meta-states.

Since $(s,c)$ is reachable, there exists a capability sequence $\overline{c}^\star=(c_1,\ldots,c_n,c)$ such that, after $c_1,\ldots,c_n$, the resulting meta-state $s^\dagger$ represents distributions $\rho_0$ and $\rho_1$ with $s\in\operatorname{supp}(\rho_0)\cup\operatorname{supp}(\rho_1)$. Thus, executing $\overline{c}^\star$ has non-zero probability of executing $c$ in $s$.

Suppose, for contradiction, that after some iteration $k$, MCQS never again produces such a query. Then $\mathcal{D}_{SC}(s,c)$ remains finite. Let $\rho=(\rho_0+\rho_1)/2$. Since $\rho(s)>0$ and the disagreement reward is non-negative, the reward of $\overline{c}^\star$ is bounded below by its revisit reward, $R(\overline{c}^\star)\geq\frac{\xi\rho(s)}{\mathcal{D}_{SC}+1}=:\beta>0$.

Now consider the queries selected after iteration $k$. By assumption, all avoid executing $c$ in $s$. Repeated selection of any such query increases the visitation counts of its encountered state-capability pairs, driving its revisitation reward to zero. Once all the finite combinations of state-capability pairs have been tried, its disagreement reward will also be zero by Lemma~\ref{lemma:reward}. Hence, its total reward converges to zero under repeated selection.

Because $P$ and $O$ are finite and queries have finite horizon, there are finitely many candidate queries. Therefore, after finitely many iterations, every query avoiding $(s,c)$ has reward less than $\beta$. However, $R(\overline{c}^\star)\geq\beta$, and, by Assumption A1, MCQS selects an optimal query. It must therefore select a query with non-zero probability of executing $c$ in $s$, contradicting the assumption.

Thus, MCQS produces such a query after finitely many iterations.
\end{proof}



Now we show that any inconsistency at a state-capability pair $(s,c)$ has a non-zero probability of being reduced after finitely many observations of performing $c$ in $s$.

\begin{lemma}
\label{lemma:cont_improvement}
Let $h^{\dagger\star}$ be the true model, expressible over vocabulary $V$, and let $h^\dagger$ be the pessimistic or optimistic model constructed from $\mathcal{D}$, respectively. If $h^\dagger$ is inconsistent with $h^{\dagger\star}$ w.r.t. state-capability pair $(s,c)$, then there exists a finite $k$ such that $k$ additional executions of $c$ in $s$ have non-zero probability of reducing the inconsistency.
\end{lemma}

\begin{proof}
Observations from performing $c$ in $s$ are i.i.d.\ draws from the true effect distribution $p^\star$ at $(s,c)$. There are four possible causes for $\exists s',\Pr_{h^\dagger}(s'|s,c)\neq \Pr_{h^{\dagger\star}}(s'|s,c)$.

\emph{Case 1 (additional effect): $\Pr_{h^\dagger}(s'\mid s,c)>0$ and $\Pr_{h^{\dagger\star}}(s'\mid s,c)=0$.} Then $\langle s,c,s'\rangle \notin\mathcal{D}$, since every observed transition is realizable under $h^{\dagger\star}$. By construction, $h^\dagger_\wedge$ classifies only transitions in $\mathcal{D}$ as consistent, so the inconsistency must occur in $h^\dagger_\vee$; thus, there exists a rule $r$ assigning non-zero probability to $\eta$ with $s\models \mathrm{ocond}(r)$. By construction of the optimistic condition
$\mathrm{ocond}$, this is possible only if $s$ has not been partitioned into another rule for $c$, and hence $(s,c)$ has not been observed in $\mathcal{D}$. Therefore, a single execution of $c$ in $s$ assigns $s$ to a partition, removing $\eta$ from the effects predicted at $(s,c)$ and hence
reducing the inconsistency. 

\emph{Case 2 (missing effect): $\Pr_{h^\dagger}(s'\mid s,c)=0$ and $p^\star_\eta = \Pr_{h^{\dagger\star}}(s'\mid s,c) > 0$.} A single additional execution of $c$ in $s$ observes $\eta$ with probability $p^\star_\eta>0$, after which $h^\dagger$ assigns $\eta$ positive probability, reducing the inconsistency. 

\emph{Case 3 (incorrect condition).} If $s$ is assigned to a partition whose effect set differs from that of $(s,c)$, the resulting discrepancy is a set of effects predicted but not realizable (Case 1) or realizable but not predicted (Case 2) at $(s,c)$. Therefore, the preceding cases apply.

\emph{Case 4 (distributional error): $\Pr_{h^\dagger}(s'\mid s,c)$ and $\Pr_{h^{\dagger\star}}(s'\mid s,c)$ are both positive but unequal.}
Suppose instead $h^\dagger$'s error at $(s,c)$ is a distributional error over the $n$ effects: $e_{(s,c)} = d_{\mathrm{VD}}(\hat p_{k_0},p^\star)\frac1n\sum_\eta|\hat p_{k_0,\eta}-p^\star_\eta| > 0$, where $\hat p_{k_0}$
is the current MLE from the $k_0$ observations of $(s,c)$ in $\mathcal{D}$. Since $h^\dagger$ is given, $\hat p_{k_0}$ and hence $e_{(s,c)}$ are fixed.
Let $\hat p_{k_0+k}$ denote the empirical MLE after $k$ additional i.i.d.\ executions of $c$ in $s$. By the strong law of large numbers applied to each of the finite effects,
$\hat p_{k_0+k}\xrightarrow[k\to\infty] {\mathrm{a.s.}}p^\star$, and since $d_{\mathrm{VD}}(\cdot,p^\star)$ is continuous, $d_{\mathrm{VD}}(\hat p_{k_0+k},p^\star)\to 0$ almost surely. Almost sure
convergence implies convergence in probability, so
$q_k := \Pr\bigl(d_{\mathrm{VD}}(\hat p_{k_0+k},p^\star) < e_{(s,c)}\bigr)\to 1$ as $k\to\infty$. By the definition of convergence, taking $\delta=\tfrac12$, there exists a finite $K$ such that $q_k>\tfrac12$ for all $k\ge K$. Hence, $k=K$ additional executions suffice for $d_{\mathrm{VD}}(\hat p_{k_0+k},p^\star) < d_{\mathrm{VD}}(\hat p_{k_0},p^\star)$ with strictly positive probability


In all cases, there is a finite $k$ such that $k$ additional executions of $c$ in $s$ have non-zero probability of reducing the inconsistency in $h^\dagger$ w.r.t $(s,c)$ and $h^{\dagger\star}$.
\end{proof}

Combining Lemmas~\ref{lemma:revisiting} and~\ref{lemma:cont_improvement}, we show that any inconsistency at a reachable state-capability pair has a non-zero probability of being reduced after finitely many iterations of \APPABBREV{}.

\begin{lemma}
\label{lemma:revisit_improvement}
Let $h^{\dagger\star}$ be the true model, expressible over vocabulary $V$, and let $h^\dagger_\land$ and $h^\dagger_\lor$ be the pessimistic and optimistic models computed by \APPABBREV{} from $\mathcal{D}$, respectively. If $\xi>0$ and either $h^\dagger_\land$ or $h^\dagger_\lor$ is inconsistent with $h^{\dagger\star}$ w.r.t. a reachable state-capability pair $(s,c)$, then there exists a finite number of iterations of \APPABBREV{} after which this inconsistency has a non-zero probability of being reduced.
\end{lemma}

\begin{proof}
Suppose either $h^\dagger_\land$ or $h^\dagger_\lor$ remains inconsistent with $h^{\dagger\star}$ w.r.t. $(s,c)$. By Lemma~\ref{lemma:cont_improvement}, some finite $k$ additional observations of $c$ in $s$ has a non-zero probability of reducing this inconsistency. 

By Lemma~\ref{lemma:revisiting}, each such observation occurs after a finite number of \APPABBREV{} iterations with non-zero probability, and this continues to hold if the inconsistency persists. Applying this at most $k$ times, \APPABBREV{} collects all $k$ observations within a finite number of iterations with non-zero probability. 

Since the probability of collecting the $k$ observations is non-zero and the probability that they reduce the inconsistency is non-zero, the inconsistency at $(s,c)$ has a non-zero probability of being reduced after finitely many iterations.
\end{proof}

Having established that any error has a non-zero probability of being reduced after finitely many iterations of \APPABBREV{}, we now show that, in the limit, the learned model almost surely has variational distance 0 from the true model over reachable transitions.

\mainthm*

\begin{proof}
Since $P$ and $O$ are finite, there are finitely many reachable state-capability pairs. Suppose $h^\dagger$ remains inconsistent with $h^{\dagger\star}$ at state-capability pair $(s,c)$. The hypotheses of Lemma~\ref{lemma:revisiting} depend only on $\xi>0$ and the reachability of $(s,c)$, not on $|\mathcal{D}_{SC}(s,c)|$, so it applies afresh at each stage: after finitely many further iterations MCQS almost surely selects a
query with non-zero probability of executing $c$ in $s$, and by Lemma\,\ref{lemma:revisit_improvement}, the resulting observations have non-zero probability of reducing the inconsistency. Applying this successively yields $|\mathcal{D}_{SC}(s,c)|\to\infty$ unless $(s,c)$ becomes consistent.

For infinitely many observations of $(s,c)$, the MLE effect distribution converges almost surely to $p^\star$ by the strong law of large numbers, and by Lemma~\ref{lemma:cont_improvement} each structural error is corrected with fixed non-zero probability per observation, hence almost surely after finitely many. The error at every reachable state-capability pair therefore converges almost surely to zero, and as there are finitely many such pairs this holds simultaneously for all of them almost surely. Since $\mathrm{VD}(h^\dagger_i,h^{\dagger\star})$ is the average of the absolute differences over the finitely many reachable transitions, each summand converging almost surely to zero gives $\mathrm{VD}(h^\dagger_i,h^{\dagger\star})\xrightarrow[i\to\infty]{\mathrm{a.s.}}0$.
\end{proof}

\subsection{Finite Transitions to Structural Convergence}

We now prove Theorem\,\ref{theorem:finite_transition_diff} introduced in Section\,\ref{sec:theory}, which states that there exists a finite number of transitions that need to be collected to make the pessimistic and optimistic models equivalent.


\supportthm*

\begin{proof}
Since $P$ and $O$ are finite, the set of states and capabilities are finite, and hence so is the set of state-capability-state transitions. Therefore, $T_{h^{\dagger\star}}$ is also finite.  

Let $T_{\mathcal{D}}$ denote the unique set of transitions observed in $\mathcal{D}$. Since every observed transition must be realizable by the true model, $T_{\mathcal{D}}\subseteq T_{h^{\dagger\star}}$. Because $ T_{h^{\dagger\star}}$ is finite, the set of unobserved transitions $T_\delta= T_{h^{\dagger\star}}\setminus T_{\mathcal{D}}$ is also finite.

Let $\mathcal{D}'=\mathcal{D}\uplus T_\delta$ denote the updated dataset with all transitions in $T_\delta$ observed. This results in $\mathcal{D}'$ having the same set of unique transitions as the true model $T_{\mathcal{D}'}=T_{\mathcal{D}}\cup T_\delta= T_{h^{\dagger\star}}$. Let $h^\dagger_\land$ and $h^\dagger_\lor$ be the pessimistic and optimistic models constructed from $\mathcal{D}'$, respectively.

By construction, $h^\dagger_\land$ admits exactly the observed transitions so $T_{h^\dagger_\land}=T_{D'}=T_{h^{\dagger\star}}$. Since $h^\dagger_\land$ is possibly sound and $h^\dagger_\lor$ is the join of all possibly sound hypotheses, $T_{h^{\dagger\star}}=T_{h^\dagger_\land}\subseteq T_{h^\dagger_\lor}$.

Suppose, for contradiction, that there exists a transition $\langle s,c,s'\rangle\in T_{h^\dagger_\lor}\setminus T_{h^{\dagger\star}}$. Since $h^\dagger_\lor$ is possibly sound, this can only occur when $(s,c)$ is unobserved in $\mathcal{D}$. However, it is always valid to ask an agent to achieve an intent, so every state--capability pair admits at least one transition in $T_{h^{\dagger\star}}$. Since $T_{\mathcal{D}'}=T_{h^{\dagger\star}}$, that transition is observed in $\mathcal{D}'$, contradicting the assumption that $(s,c)$ is unobserved.

Therefore, $T_{h_\lor'}= T_{h^{\dagger\star}}=T_{h_\land'}$, and thus $h_\land'\equiv h_\lor'$.
\end{proof}

\subsection{Convergence of Consistent Models}

We now prove that, when $\mathcal{D}$ is transition and effect complete, if $h^\dagger_\land\equiv h^\dagger_\lor$ then (1) both $h^\dagger_\land$ and $ h^\dagger_\lor$ are equivalent to $h^{\dagger\star}$; (2) a model $h^\dagger$ that is complete and possibly sound w.r.t. $\mathcal{D}$ is also equivalent to $h^{\dagger\star}$. We first prove that the pessimistic and optimistic models are equivalent to the true model when these conditions hold.

\begin{theorem}
\label{theorem:termination}
Let $h^{\dagger\star}$ be the true agent capability model, and assume $h^{\dagger\star}$ is expressible over vocabulary $V$. Let $\mathcal{D}$ be transition and effect complete. Then, at any stage of \APPABBREV{}, if $h^\dagger_\land\equiv h^\dagger_\lor$, it follows that $h_\land\equiv h^\dagger_\lor\equiv h^{\dagger\star}$.
\end{theorem}

\begin{proof}
Since $h^{\dagger\star}$ is expressible in $V$, $T_{h^{\dagger\star}}$ is the consistent.

By the premises, $h^\dagger_\land\equiv h^\dagger_\lor$ meaning $T_{h^\dagger_\land}=T_{h^\dagger_\lor}=T_c$. Suppose, for contradiction, that $T_c\neq T_{h^{\dagger\star}}$. Then either there exists a transition $\langle s,c,s'\rangle\in T_{h^{\dagger\star}}\setminus T_c$, or there exists a transition $\langle s,c,s'\rangle\in T_c\setminus T_{h^{\dagger\star}}$.

First, consider $\langle s,c,s'\rangle\in T_{h^{\dagger\star}}\setminus T_c$. If $(s,c)$ had previously been observed, then, by the premise that $\mathcal{D}$ is transition complete, all possible effects of executing $c$ in $s$ would have been sampled. Since $\langle s,c,s'\rangle\in T_{h^{\dagger\star}}$, the transition would therefore have been observed, contradicting $\langle s,c,s'\rangle\notin T_c$. Hence, $(s,c)$ must be unobserved.

By the premises that $\mathcal{D}$ is effect complete, there exists a $\langle s_1,c,s_2\rangle\in\mathcal{D}$ s.t. $\eta(s,s')=\eta(s_1,s_2)$. By construction of the optimistic model, there must exist a rule $r$ s.t. $\eta(s,s')\in\mathrm{effects}(r)$. Since $(s,c)$ is unobserved, it cannot be assigned to any rule in the optimistic model. Therefore, by the construction of the optimistic model, $s_1\models\mathrm{ocond}(r)$, where $\mathrm{ocond}(r)$ is the optimistic condition of $r$. Therefore, $\langle s,c,s'\rangle$ is consistent under $h^\dagger_\lor$. In contrast, $h^\dagger_\land$ classifies only observed transitions as consistent and therefore does not classify $\langle s,c,s'\rangle$ as consistent. Thus, $h^\dagger_\land\not\equiv h^\dagger_\lor$, yielding a contradiction.

Now consider $\langle s,c,s'\rangle\in T_c\setminus T_{h^{\dagger\star}}$. By construction, $h^\dagger_\land$ classifies only observed transitions as consistent, so $\langle s,c,s'\rangle$ must have been observed. Every observed transition is realizable under the true model, implying $\langle s,c,s'\rangle\in T_{h^{\dagger\star}}$, again yielding a contradiction.

Therefore, $T_c=T_{h^{\dagger\star}}$, and hence $h^\dagger_\land\equiv h^\dagger_\lor\equiv h^{\dagger\star}$.
\end{proof}

Due to the pessimistic and optimistic models bounding the lattice of consistent hypotheses when $\mathcal{D}$ is transition and effect complete, we can now prove that any other consistent hypothesis must also be equivalent to the true model when the pessimistic and optimistic models are equivalent.

\begin{theorem}
\label{theorem:all_sc_model}
Let $h^\dagger$ be any model of the discovered capabilities $C$ that is complete and possibly sound with respect to the dataset $\mathcal{D}$, collected during a run of \APPABBREV, in which $\mathcal{D}$ is transition and effect complete. If the true model $h^{\dagger\star}$ is expressible over vocabulary $V$, and the pessimistic and optimistic models $h^\dagger_\land$ and $h^\dagger_\lor$ learned from $\mathcal{D}$ using \APPABBREV{} satisfy $h^\dagger_\land\equiv h^\dagger_\lor$, then $h^\dagger_\land\equiv h^\dagger_\lor\equiv h^\dagger\equiv h^{\dagger\star}$.
\end{theorem}

\begin{proof}
Since $h^\dagger$ is complete with respect to $\mathcal{D}$, and $h_\land$ is the meet of all complete hypotheses, we have $h_\land\preceq h^\dagger$. Additionally, since $h^\dagger$ is possibly sound with respect to $\mathcal{D}$, and $h_\lor$ is the join of all possibly sound hypotheses, we have $h^\dagger\preceq h_\lor$.

When $h_\land\equiv h_\lor$, these inequalities imply
$h_\land\equiv h^\dagger\equiv h_\lor$.

By Theorem~\ref{theorem:termination}, convergence of the pessimistic and optimistic models when $\mathcal{D}$ is transition and effect complete implies that both coincide with the true model. Therefore,
$h_\land\equiv h_\lor\equiv h^\dagger\equiv h^{\dagger\star}$.
\end{proof}


\section{Comparison With Action Model Learning}
\label{sec:comp_action_model_learning}

\begin{table*}[]
    \centering
    \begin{small}
    \begin{tabular}{r|cc}
        & Action Model Learning & Capability Learning \\
        Works without input action headers & {\color{OrangeRed}\ding{55}} & {\color{ForestGreen}\ding{52}} \\
        Discovers and predicts agent capabilities & {\color{OrangeRed}\ding{55}} & {\color{ForestGreen}\ding{52}} \\
        Models  capture long-horizon execution & {\color{OrangeRed}\ding{55}} & {\color{ForestGreen}\ding{52}} \\
        Observation and modeling over different vocabularies  & {\color{OrangeRed}\ding{55}} & {\color{ForestGreen}\ding{52}} \\
        Evaluates planning and reasoning capabilities & {\color{OrangeRed}\ding{55}} & {\color{ForestGreen}\ding{52}} \\
        Models feature arbitrary condition-effect rules &  {\color{OrangeRed}\ding{55}} & {\color{ForestGreen}\ding{52}} \\
        Models primitive actions of agents & {\color{ForestGreen}\ding{52}} & {\color{OrangeRed}\ding{55  }}\\
        
        \\
    \end{tabular}
    \caption{Key differences between traditional action model learning and learning intent-driven capabilities.}
    \label{tab:diff_bet_action_and_capability_learning}
    \end{small}
\end{table*}

As discussed in the Introduction, action model learning and capability model learning differ in several key ways, summarized in Table~\ref{tab:diff_bet_action_and_capability_learning}. Action model learning assumes access to, or seeks to recover, a model of the agent's primitive actions. In contrast, capability model learning does not require knowing the agent's action set, nor does it require observing and modeling the agent's behavior in the environment's vocabulary. Instead, it characterizes the long-horizon outcomes the agent can bring about through its planning and reasoning. Thus, action model learning models the effects of actions available in an environment, whereas capability model learning models what a particular agent is capable of achieving.

\section{Implementation of Compact Distributions Over States}
\label{sec:exact_implementation_details}

This section provides implementation details regarding our representation of a distribution over states. For illustration purposes, let $h_c$ be a hypothesis about capability $c$, $r \in h_c$ be a conditional effect rule, $\textrm{cond}(r)$ be the well-formed formula corresponding to the condition for $r$, and $\textrm{effects}(r)=\{(p_i,\eta_i)\}$ be the set of probabilistic effects of $r$. $p_i$ is the probability of $\eta_i$ occurring, and $\eta_i$ is a conjunction of problem literals.

\paragraph{States as Bit Vectors}
We represent a problem state $x$ as a binary vector, where the $j$'th bit corresponds to the truth value of the $j$'th problem literal assignment of $x$ according to some fixed ordering. We represent a distribution over possible states $S$ using a hash map $S: x \rightarrow \log P(x)$: keys are states $x$ in bit vector form, and values are the log-probability mass associated with the state $\log P(x)$. This representation is sparse; we only store states with probability mass greater than zero so as not to materialize the exponential state space when possible. 

\paragraph{Capability Conditions as DNFs}
We represent a capability condition $\textrm{cond}(r)$ as either a Disjunctive Normal Form (DNF) formula or the negation of one. Each clause $\textrm{clause}_k \in \textrm{cond}(r)$ is a conjunction of literals stored as a bit vector, similarly to how states are stored. This enables efficient checking of whether a state satisfies a capability condition via efficient bitwise operations.

\paragraph{Effects as Bit Vectors}
Similar to how we represent states, we represent probabilistic effect outcomes $\eta_i$ as bit vectors, with each bit corresponding to a literal truth value. Furthermore, we store a binary mask $\textrm{mask}_i$ that is $1$ in position $k$ when $\eta_i$ affects literal $k$ and $0$ otherwise.

\paragraph{State Distribution Updates}
Let $x \models \textrm{cond}(r)$ denote that $x$ is a model of $\textrm{cond}(r)$, and $S_1+S_2=S$ denote the distribution obtained by taking the sum of distributions $S_1$ and $S_2$. Utilizing this efficient satisfiability check, we update a state distribution using the following algorithm:

\begin{algorithm}[h]
\caption{Update State Distribution}
\label{alg:statedistupdate}
\begin{algorithmic}[1]
\STATE \textbf{Inputs:} state distribution $S$, conditional effect $r$
\STATE \textbf{Output:} new state distribution $S'$
\STATE $S_{change} \leftarrow \{(x_k \rightarrow \log P(x_k)) | x_k \models \textrm{cond}(r)\}$ 
\STATE $S_{nochange} \leftarrow \{(x_k \rightarrow \log P(x_k)) | x_k \not \models \textrm{cond}(r)\}$ 
\STATE $S' \leftarrow S_{nochange}$
\FOR{$(p_i, \eta_i, \textrm{mask}_i) \in \textrm{effects}(r)$}
    \STATE $S_i \gets \{((x_k \land \textrm{mask}_i) \lor \eta_i \rightarrow \log [P(x_k) * p_i]) | (x_i \rightarrow \log P(x_i)) \in S_{change}\}$
    \STATE $S' \gets S' + S_i$
\ENDFOR
\RETURN $S'$
\end{algorithmic}
\end{algorithm}

\section{Additional Agent and Environment Details}
\label{sec:agent_env_details}

In this section, we go through each evaluated agent and environment, and provide additional details on how they were implemented.

\subsection{MiniGrid}

\subsubsection{Environment}

\begin{figure}[H]
    \centering
    \includegraphics[width=0.5\linewidth]{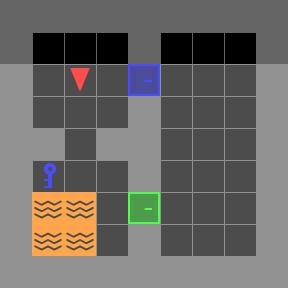}
    \caption{Minigrid environment evaluated}
    \label{fig:minigrid}
\end{figure}

We use a $9\times9$ MiniGrid (shown in Fig.\,\ref{fig:minigrid}) that contains walls, lava, a blue key, a locked green door, and a locked blue door. The grid is divided into 4 quadrants, with a single-tile path from each quadrant to each adjacent one. The blue door is between the northwest and northeast quadrants, and the green door is between the southwest and southeast quadrants. The blue key is located in the southwest quadrant near the lava. Finally, the agent starts in the northwest quadrant but cannot drop the key.

Therefore, there are some interesting capabilities in this domain. The agent can traverse near the lava to pickup the blue key, use the blue key to unlock the blue door, open/close the blue door, and traverse to any of the locations. 

\subsubsection{Symbolic Representation}

As mentioned, instead of using (x,y) locations in the grid, the locations of the key and agent are tracked based on their quadrant. For capabilities, a location in the quadrant is sampled to give as the intent to the agent.

\begin{table}[H]
\centering
\begin{tabular}{p{0.45\linewidth} p{0.45\linewidth}}
\toprule
Predicate & Description \\
\midrule
\texttt{agent-at(q)} & The agent is in quadrant q \\
\texttt{blue-key-at(q)} & The blue key is in quadrant q \\
\texttt{carrying-blue-key()} & The agent is carrying the blue key \\
\texttt{is-dead()} & The agent is dead \\
\texttt{green-door-open()} & The green door is open \\
\texttt{green-door-locked()} & The green door is locked \\
\texttt{green-door-closed()} & The green door is closed \\
\texttt{blue-door-open()} & The blue door is open \\
\texttt{blue-door-locked()} & The blue door is locked \\
\texttt{blue-door-closed()} & The blue door is closed \\
\bottomrule
\end{tabular}
\caption {Predicate vocabulary used in MiniGrid. }
\end{table}

\subsubsection{ReAct Agent}

The ReAct agent is designed to describe the current state in text, prompt the agent to achieve the objective, and re-plan if necessary. 

The base prompt given to the LLM uses the height, width, and action names, specifies the actions and rules, and gives a couple of examples.

\begin{verbatim}
"You are navigating a {height}x{width}
grid. Rows 1-{height} increase going 
south; columns 1-{width} increase going
east. Coordinate format: (row, col).

GRID LEGEND
===========
 .  = empty cell
 W  = wall (impassable)
 L  = lava (instant death -- NEVER step
 here)
bK  = blue key  (first letter = color)
bD  = blue door (locked/closed)
b=  = blue door (open)
gD  = green door (locked/closed)
g=  = green door (open)
A>  = you, facing east  (^ north, 
v south, < west, > east)

ACTIONS
=======
{action_names}

- face_north / face_south / face_east / 
face_west : rotate to face that 
compass direction (no movement).
- move_forward : move one cell in the
direction you face. Blocked by walls,
closed/locked doors, objects, and lava.
- pickup : pick up the object in the
cell you face (you must be adjacent 
and facing it). You can carry only one
object.
- toggle : open or close the door in
the cell you face. Locked doors 
require holding the matching color key.
- done : signal that the goal is
achieved or no further action is
possible.

RULES
=====
Movement:
- move_forward moves you one cell in
the direction you face.
- You CANNOT move into walls (W), 
lava (L), closed/locked doors, or cells 
with objects.
- Lava kills instantly -- NEVER step 
into a lava cell.
- You cannot move into a cell occupied 
by an object; pick it up from an 
adjacent cell instead.

Interaction:
- You must be adjacent to and facing a
cell to interact with it (pickup or 
toggle).
- You can carry at most one object at a 
time.

Doors and keys:
- Doors block movement when closed or 
locked, just like walls.
- A locked door is opened by holding 
the matching COLOR key and using 
toggle. A blue key opens a blue door, 
a green key opens a green door, etc.
- Once opened, a door becomes a 
passable cell -- you can walk through 
it freely.
- IMPORTANT: doors are the ONLY way 
to cross through wall barriers. Before 
concluding a goal is unreachable, 
check every door on the map and whether 
you have (or can get) the matching key.

Planning:
- Before acting, scan the grid for 
barriers (walls, lava) and identify a
all possible paths including through 
doors.
- For each move_forward in your plan, 
verify the target cell is empty or an 
open door.
- When the goal is on the other side 
of a wall, find a door that crosses 
the barrier and plan how to unlock it.

EX 1 -- Navigate and pick up an object
======================================
Observation:
     1  2  3  4  5
  1  .  .  .  .  .
  2  . Av  .  .  .
  3  .  .  . bK  .
  4  .  .  W  W  .
  5  .  .  .  .  .
Agent: (2, 2) facing south
Carrying: nothing

Goal: Pick up the blue key.

Thought:
I am at (2, 2) facing south. The blue 
key is at (3, 4).
Plan a path: go south one step to (3,
2), then east toward column 4.
- move_forward: I face south, so I 
move to (3, 2). Cell is empty -- safe.
- face_east: I now face east at (3, 2).
- move_forward: move to (3, 3). Cell is
empty -- safe.
- move_forward: (3, 4) has the blue key
-- BLOCKED, I cannot move into an 
occupied cell.
  Instead I stay at (3, 3) facing east,
  and the key is in the cell I face.
- pickup: pick up the blue key from
(3, 4).

Actions:
move_forward
face_east
move_forward
pickup

EXAMPLE 2 -- Unlock a door
===========================
Observation:
     1  2  3  4  5
  1  .  .  .  .  .
  2  .  .  A>  .  .
  3  .  .  W bD  .
  4  .  .  .  .  .
  5  .  .  .  .  .
Agent: (2, 3) facing east
Carrying: a blue key

Goal: Open the blue door.

Thought:
I am at (2, 3) facing east, carrying
the blue key. The blue door is at
(3, 4).
I need to be adjacent to the door and
facing it. From (2, 4) I could face 
south toward (3, 4).
- move_forward: move east to (2, 4).
Cell is empty -- safe.
- face_south: now I face south at
(2, 4). The cell ahead is (3, 4) which
is the blue door.
- toggle: I hold the blue key so the
locked blue door unlocks and opens.

Actions:
move_forward
face_south
toggle

OUTPUT FORMAT
=============
Always output exactly:
Thought: <your step-by-step reasoning
-- trace each action, verify the 
target cell is safe>
Actions:
<one action per line from the action
list above, max {max_plan_length} 
actions>
"
\end{verbatim}

This prompt is used as the system prompt, followed by a description of the task and the state. It then parses out the actions to take. In the case of an invalid action, the following prompt is provided.

\begin{verbatim}
"I could not parse your previous 
response. Please reply again using
exactly the required format:

Thought: <your reasoning>
Actions:
<one action per line>

Valid actions: {action_names}

Each line under Actions: must contain
exactly one action name from the
list above, nothing else. If the goal
is already achieved or impossible,
output a single action: done"
\end{verbatim}

Then the policy is executed until either the maximum number of actions is reached or the LLM returns the ``done'' action.

\subsection{Overcooked}

\subsubsection{Environment}

\begin{figure}[H]
    \centering
    \includegraphics[width=0.5\linewidth]{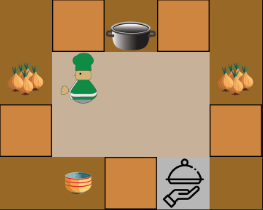}
    \caption{Overcooked environment evaluated.}
    \label{fig:overcooked}
\end{figure}

We use Overcooked, a cooperative kitchen gridworld environment. At each timestep, the agent selects an action from a discrete action space, thereby updating the environment state and producing a new observation.

The agent must navigate, collect ingredients, prepare dishes, and deliver them. The task involves a multi-step process: collect the onion, add it to the pot, wait for it to cook, retrieve the soup, and deliver. The environment includes stochastic elements, namely a 5-stage cooking process.

We use a single predefined initial state (shown in Fig.\,\ref{fig:overcooked}) that specifies the kitchen layout, the agent's position, the ingredient pile locations, the pot placement, and the delivery counter.

\begin{table}[H]
\centering
\begin{tabular}{p{0.28\linewidth} p{0.62\linewidth}}
\toprule
\textbf{Attribute} & \textbf{Description} \\
\midrule
kitchen\_layout & Fixed configuration of counters and tiles \\
agent\_position & Initial position of the agent in the grid \\
ingredient\_piles & Location of onion piles \\
pot\_location & Placement of cooking pot \\
delivery\_counter & Location of delivery area \\
\bottomrule
\end{tabular}
\caption {Attributes defining the predefined initial state.}
\end{table}

\subsubsection{Symbolic Representation}

We construct a symbolic state representation that maps low-level environment states to a set of logical predicates. The representation operates on the structured state exposed by the Overcooked environment, including agent location, held items, pot state, and object positions.

Given a state, the symbolic representation provides a set of ground literals over a fixed predicate vocabulary. Objects are limited to the agent (chef1), items (onion, soup-dish), locations (onion-pile-1, onion-pile-2, pot, delivery), and pot cooking stages.

Each predicate is computed directly from the underlying state. Position predicates (e.g., \texttt{at}) are obtained by querying the agent's location, while state predicates (e.g., \texttt{pot-state}, \texttt{holding}) are determined from the environment's internal state representation.

\begin{table}[H]
\centering
\begin{tabular}{p{0.40\linewidth} p{0.50\linewidth}}
\toprule
Predicate & Description \\
\midrule
\texttt{at(x, location)} & Entity x is at location \\
\texttt{holding(agent, x)} & Agent possesses item x \\
\texttt{in-pot(x)} & Item x is in the pot \\
\texttt{pot-state(state)} & Current cooking stage \\
\texttt{clean(dish, location)} & Clean dish at location \\
\texttt{at-pile(onion, pile)} & Onion at specified pile \\
\texttt{delivered(x)} & Item x has been delivered \\
\bottomrule
\end{tabular}
\caption {Predicate vocabulary used in the symbolic representation.}
\end{table}

\subsubsection{HDDLGym Agent}

We use an agent trained using hierarchical reinforcement learning from HDDLGym \citep{La_Mon-Williams_Shah_2025}. The agent operates in single-agent mode in the Overcooked environment.

The agent receives high-level capability intents and executes learned low-level policies to achieve these intents. Our experiments show that spurious correlations in training data can induce unintended behaviors.

MCQS discovered 5 capabilities: get onion (c-0), add onion to pot (c-1), cooking progression (c-2), retrieve soup (c-3), and deliver soup (c-4). These capabilities form a linear dependency chain where each is the sole producer of its output state. The cooking capability (c-2) is the only temporal capability with length 5, modeling the stochastic multi-stage cooking process with probabilistic progression through cooking stages.

\subsection{Crafter}

\subsubsection{Environment}

\begin{figure}[H]
    \centering
    \includegraphics[width=0.5\linewidth]{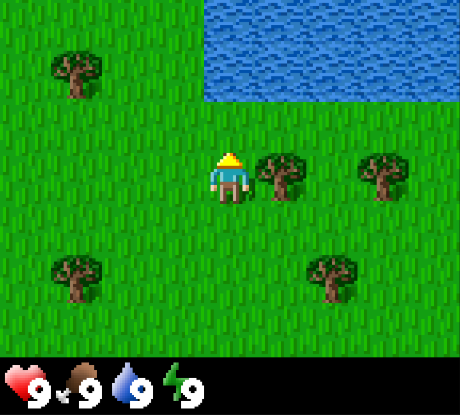}
    \caption{Crafter environment evaluated.}
    \label{fig:crafter}
\end{figure}

We use Crafter, a grid-based survival and crafting environment. At each timestep, the agent selects an action from a discrete action space, thereby updating the environment state and producing a new observation.

The agent must collect resources, craft tools, and interact with terrain and objects. Reward is given based on the achievements unlocked by the agent; there are 22 such achievements in the Crafter world.

We use a fixed $9\times 9$ grid initial state (shown in Fig.\,\ref{fig:crafter}) containing grass, tree, and water locations with fixed stats for the player. This state was specified using the structured state representation provided by Crafter-OO \citep{hafner2022benchmarking,khan2025one}. 

\begin{table}[H]
\centering
\begin{tabular}{p{0.34\linewidth} p{0.56\linewidth}}
\toprule
Attribute & Description \\
\midrule
\texttt{area} & Size of the environment grid \\
\texttt{player\_position} & Initial position of the agent in the grid \\
\texttt{player\_facing} & Initial orientation of the agent \\
\texttt{terrain\_layout} & Fixed terrain configuration of the environment \\
\texttt{objects} & Initial placement of objects in the environment \\
\bottomrule
\end{tabular}
\caption{Attributes defining the predefined initial states.}
\end{table}

\subsubsection{Symbolic Representation}

We construct a symbolic state representation that maps low-level environment states to a set of logical predicates. The representation operates on the structured state exposed by Crafter-OO, including terrain, objects, player attributes, inventory, and achievements.

Given a state, the symbolic representation provides a set of ground literals over a fixed predicate vocabulary. Objects are limited to a predefined set of obstacles, tools, structures, and achievements. This symbolic state is not a perfect representation of the underlying environment state, as it retains only the information necessary for decision-making and removes irrelevant details.

Each predicate is computed directly from the underlying state. Some predicates are related to position (e.g., \texttt{next\_to}, \texttt{close\_to}), which are obtained by calculating the Manhattan distances between the agent and nearby objects, while other predicates (e.g., \texttt{has}, \texttt{can\_make}, \texttt{can\_build}) are determined from inventory contents.

We specify the goal as a symbolic literal conjunction over this set of predicates.

\begin{table}[H]
\centering
\begin{tabular}{p{0.28\linewidth} p{0.62\linewidth}}
\toprule
Predicate & Description \\
\midrule
\texttt{next\_to(x)} & Agent is adjacent to object $x$ \\
\texttt{close\_to(x)} & Agent is within a Manhattan distance of 3 from $x$ \\
\texttt{exists(x)} & Object $x$ exists in the environment \\
\texttt{is\_sleeping()} & Agent is sleeping \\
\texttt{has(x)} & Agent possesses tool $x$ \\
\texttt{can\_make(x)} & Agent has resources to craft $x$ \\
\texttt{can\_build(x)} & Agent has resources to build $x$ \\
\texttt{achieved(x)} & Achievement $x$ has been completed \\
\bottomrule
\end{tabular}
\caption{Predicate vocabulary used in the symbolic representation.}
\end{table}

\subsubsection{Qwen-Steve Agent}

We use an agent that combines the Qwen2.5-7B-Instruct \citep{qwen2.5} model with Crafter Steve1 from CrafterDojo to execute skills. Steve-1 executes higher-level skills as sequences of low-level Crafter actions.

The agent operates over a fixed subset of skills derived from the CrafterDojo skill set. CrafterDojo defines 61 skills; we use a subset of 16.

\begin{multicols}{2}
\begin{itemize}
\item \texttt{craft wood pickaxe}
\item \texttt{craft wood sword}
\item \texttt{dig a tunnel}
\item \texttt{go explore}
\item \texttt{go to sleep}
\item \texttt{move to east}
\item \texttt{move to north}
\item \texttt{move to south}
\item \texttt{move to west}
\item \texttt{obtain coal}
\item \texttt{obtain diamond}
\item \texttt{obtain tree}
\item \texttt{obtain water}
\item \texttt{place crafting table on grass}
\item \texttt{place crafting table to build shelter}
\item \texttt{stay}
\end{itemize}
\end{multicols}

The agent takes a symbolic goal and converts it into task text. At each step, the current environment state is mapped to a symbolic state representation and then converted into state text. The task text and state text are passed to the Qwen model to select a skill.

Given the task text, state text, and candidate skills, the agent uses Qwen to select the next skill to execute. This selection is repeated after each skill attempt, allowing the agent to adapt based on the updated environment state.

The selected skill is executed by CSteve1 for multiple timesteps. We use greedy action selection during skill execution, making CSteve1’s behavior deterministic. After each environment step, the agent updates the symbolic state and checks whether the original symbolic goal is satisfied. If the goal is achieved, the agent returns success immediately. If not, CSteve1 continues executing the current skill until the fixed step limit for that skill is reached.

After a skill attempt finishes, if the symbolic goal is still not achieved, the agent selects a skill again using the updated state and the same original goal. The number of skill selection attempts is fixed. If the goal is still not achieved after that limit, the agent terminates and returns failure.

\subsubsection{Skill Selection}

We score a set of candidate skills using the LLM to select the skill to be executed. We convert the goal and current-state representations into text, then evaluate each skill independently.

For each skill, we prompt the Qwen model using the goal description and state text, along with the skill name, and ask whether the skill is the best to execute to achieve the goal from the current state. We calculate two log-probabilities: $\log P(\text{yes})$ --- the probability of the model answering yes for the prompt, and $\log P(\text{no})$ --- the probability of the model answering no for the prompt. We calculate the difference between $\log P(\text{yes})$ and $\log P(\text{no})$ to score each individual skill. A high score means the model thinks the skill is more suitable for execution than not. We then pick the skill with the highest score.

An example prompt used for scoring a candidate skill is:

\begin{verbatim}
  " Goal text: [goal description]
    Current state: [state description]
    Candidate skill: [skill name]
    Is this candidate the best skill to
    execute next for the goal?
    Answer yes or no."
\end{verbatim}

\subsection{PDDLGym}

PDDLGym~\citep{silver2020pddlgym} is a benchmark suite of planning environments based on the Planning Domain Definition Language (PDDL). In PDDLGym, each environment is specified by a PDDL domain file and a PDDL problem file. The domain file defines the predicate vocabulary and object types, while the problem file defines the objects, initial state, and goal condition. We use five PDDLGym environments: Blocksworld, Depot, First Responders, Tireworld, and Probabilistic Elevators.

\subsubsection{Blocksworld}

Blocksworld is a block-stacking environment in which blocks are rearranged on a table using a robot hand.

\begin{table}[H]
\centering
\small
\begin{tabular}{p{0.30\linewidth} p{0.60\linewidth}}
\toprule
\textbf{Predicate} & \textbf{Meaning} \\
\midrule
\texttt{on(x, y)} & Block \texttt{x} is on block \texttt{y}. \\
\texttt{ontable(x)} & Block \texttt{x} is on the table. \\
\texttt{clear(x)} & Block \texttt{x} has no block on top of it. \\
\texttt{handempty(r)} & Robot \texttt{r} is not holding a block. \\
\texttt{handfull(r)} & Robot \texttt{r} is holding a block. \\
\texttt{holding(x)} & Block \texttt{x} is being held. \\
\bottomrule
\end{tabular}
\caption{Predicates for Blocksworld.}
\label{tab:blocksworld-predicates}
\end{table}

\begin{table}[H]
\centering
\small
\begin{tabular}{p{0.22\linewidth} p{0.68\linewidth}}
\toprule
\textbf{Component} & \textbf{Specification} \\
\midrule
Objects &
\texttt{a}, \texttt{b}, and \texttt{c} of type \texttt{block}; \texttt{robot} of type \texttt{robot}. \\
Initial state &
\texttt{clear(a)}, \texttt{clear(b)}, \texttt{clear(c)}, \texttt{ontable(a)}, \texttt{ontable(b)}, \texttt{ontable(c)}, and \texttt{handempty(robot)}. \\
\bottomrule
\end{tabular}
\caption{Problem instance used for Blocksworld.}
\label{tab:blocksworld-problem}
\end{table}

\subsubsection{First Responders}

First Responders is an emergency-response environment involving fire units, medical units, victims, fires, hospitals, and water sources.

\begin{table}[H]
\centering
\small
\begin{tabular}{p{0.45\linewidth} p{0.45\linewidth}}
\toprule
\textbf{Predicate} & \textbf{Meaning} \\
\midrule
\texttt{fire(l)} & Location \texttt{l} has a fire. \\
\texttt{nfire(l)} & Location \texttt{l} does not have a fire. \\
\texttt{victim-at(v, l)} & Victim \texttt{v} is at location \texttt{l}. \\
\texttt{victim-healthy(v)} & Victim \texttt{v} is healthy. \\
\texttt{victim-hurt(v)} & Victim \texttt{v} is hurt. \\
\texttt{victim-dying(v)} & Victim \texttt{v} is dying. \\
\texttt{hospital(l)} & Location \texttt{l} is a hospital. \\
\texttt{water-at(l)} & Water is available at location \texttt{l}. \\
\texttt{adjacent(l1, l2)} & Location \texttt{l1} is adjacent to location \texttt{l2}. \\
\texttt{fire-unit-at(u, l)} & Fire unit \texttt{u} is at location \texttt{l}. \\
\texttt{medical-unit-at(u, l)} & Medical unit \texttt{u} is at location \texttt{l}. \\
\texttt{have-water(u)} & Fire unit \texttt{u} has water. \\
\texttt{have-victim-in-unit(v, u)} & Victim \texttt{v} is inside medical unit \texttt{u}. \\
\bottomrule
\end{tabular}
\caption{Predicates for First Responders.}
\label{tab:first-responders-predicates}
\end{table}

\begin{table}[H]
\centering
\small
\begin{tabular}{p{0.22\linewidth} p{0.68\linewidth}}
\toprule
\textbf{Component} & \textbf{Specification} \\
\midrule
Objects &
\texttt{l1} and \texttt{l2} of type \texttt{location}; \texttt{f1} of type \texttt{fire\_unit}; \texttt{m1} of type \texttt{medical\_unit}; \texttt{z1} and \texttt{z2} of type \texttt{victim}. \\
Initial state &
\texttt{hospital(l2)}, \texttt{water-at(l1)}, \texttt{fire(l1)}, \texttt{fire(l2)}, \texttt{victim-at(z1, l2)}, \texttt{victim-hurt(z1)}, \texttt{victim-at(z2, l2)}, \texttt{victim-dying(z2)}, \texttt{adjacent(l1, l2)}, \texttt{adjacent(l2, l1)}, \texttt{fire-unit-at(f1, l1)}, and \texttt{medical-unit-at(m1, l2)}. \\
\bottomrule
\end{tabular}
\caption{Problem instance used for First Responders.}
\label{tab:first-responders-problem}
\end{table}

\subsubsection{Tireworld}

Tireworld is a navigation environment in which a vehicle moves between locations and may need to change a flat tire using spare tires.

\begin{table}[H]
\centering
\small
\begin{tabular}{p{0.34\linewidth} p{0.56\linewidth}}
\toprule
\textbf{Predicate} & \textbf{Meaning} \\
\midrule
\texttt{vehicle-at(loc)} & The vehicle is at location \texttt{loc}. \\
\texttt{spare-in(loc)} & A spare tire is available at location \texttt{loc}. \\
\texttt{road(from, to)} & There is a road from location \texttt{from} to location \texttt{to}. \\
\texttt{not-flattire()} & The vehicle does not currently have a flat tire. \\
\bottomrule
\end{tabular}
\caption{Predicates for Tireworld.}
\label{tab:tireworld-predicates}
\end{table}

\begin{table}[H]
\centering
\small
\begin{tabular}{p{0.22\linewidth} p{0.68\linewidth}}
\toprule
\textbf{Component} & \textbf{Specification} \\
\midrule
Objects &
\texttt{l-1-1}, \texttt{l-1-2}, \texttt{l-1-3}, \texttt{l-2-1}, \texttt{l-2-2}, and \texttt{l-3-1} of type \texttt{location}. \\
Initial state &
\texttt{vehicle-at(l-2-1)}, \texttt{road(l-1-1, l-1-2)}, \texttt{road(l-1-2, l-1-3)}, \texttt{road(l-1-1, l-2-1)}, \texttt{road(l-1-2, l-2-2)}, \texttt{road(l-2-1, l-1-2)}, \texttt{road(l-2-2, l-1-3)}, \texttt{road(l-2-1, l-3-1)}, \texttt{road(l-3-1, l-2-2)}, \texttt{spare-in(l-2-1)}, \texttt{spare-in(l-2-2)}, \texttt{spare-in(l-3-1)}, and \texttt{not-flattire()}. \\
\bottomrule
\end{tabular}
\caption{Problem instance used for Tireworld.}
\label{tab:tireworld-problem}
\end{table}

\subsubsection{Probabilistic Elevators}

Probabilistic Elevators is an elevator-navigation environment in which the agent moves across floors and positions, enters and exits elevators, and collects coins.

\begin{table}[H]
\centering
\small
\begin{tabular}{p{0.41\linewidth} p{0.49\linewidth}}
\toprule
\textbf{Predicate} & \textbf{Meaning} \\
\midrule
\texttt{dec\_f(f, g)} & Floor \texttt{f} is one step below floor \texttt{g}. \\
\texttt{dec\_p(p, q)} & Position \texttt{p} is one step left of position \texttt{q}. \\
\texttt{in(e, f)} & Elevator \texttt{e} is at floor \texttt{f}. \\
\texttt{at(f, p)} & The agent is at floor \texttt{f} and position \texttt{p}. \\
\texttt{shaft(e, p)} & Elevator \texttt{e} is associated with shaft position \texttt{p}. \\
\texttt{inside(e)} & The agent is inside elevator \texttt{e}. \\
\texttt{gate(f, p)} & There is a gate at floor \texttt{f} and position \texttt{p}. \\
\texttt{coin-at(c, f, p)} & Coin \texttt{c} is at floor \texttt{f} and position \texttt{p}. \\
\texttt{have(c)} & The agent has collected coin \texttt{c}. \\
\texttt{underground()} & The agent is underground. \\
\texttt{is-first-floor(f)} & Floor \texttt{f} is the first floor. \\
\texttt{is-first-position(p)} & Position \texttt{p} is the first position. \\
\bottomrule
\end{tabular}
\caption{Predicates for Probabilistic Elevators.}
\label{tab:probabilistic-elevators-predicates}
\end{table}

\begin{table}[H]
\centering
\small
\begin{tabular}{p{0.22\linewidth} p{0.68\linewidth}}
\toprule
\textbf{Component} & \textbf{Specification} \\
\midrule
Objects &
\texttt{f1}, \texttt{f2}, and \texttt{f3} of type \texttt{floor}; \texttt{p1}, \texttt{p2}, \texttt{p3}, and \texttt{p4} of type \texttt{pos}; \texttt{e1} and \texttt{e2} of type \texttt{elevator}; \texttt{c1}, \texttt{c2}, and \texttt{c3} of type \texttt{coin}. \\
Initial state &
\texttt{is-first-floor(f1)}, \texttt{is-first-position(p1)}, \texttt{underground()}, \texttt{dec\_f(f2, f1)}, \texttt{dec\_f(f3, f2)}, \texttt{dec\_p(p2, p1)}, \texttt{dec\_p(p3, p2)}, \texttt{dec\_p(p4, p3)}, \texttt{shaft(e1, p3)}, \texttt{in(e1, f1)}, \texttt{shaft(e2, p3)}, \texttt{in(e2, f1)}, \texttt{coin-at(c1, f2, p3)}, \texttt{coin-at(c2, f3, p3)}, \texttt{coin-at(c3, f1, p1)}, \texttt{gate(f2, p4)}, \texttt{gate(f3, p3)}, and \texttt{gate(f3, p4)}. \\
\bottomrule
\end{tabular}
\caption{Problem instance used for Probabilistic Elevators.}
\label{tab:probabilistic-elevators-problem}
\end{table}

\subsection{SayCan}

\subsubsection{Environment}

The SayCan agent~\citep{ahn2022saycan} is a mobile manipulator in a PyBullet simulation environment. It interacts with four blocks and one cup on a tabletop. The agent is equipped with low-level skills (e.g., pick-and-place a yellow block). It uses a controlled based on Llama-3.1-8B-Instruct \citep{meta2024llama3} to select a skill to execute based on its abstract description. \APPABBREV{} uses the same abstraction function for learning the capability model.

\subsubsection{Symbolic Representation}

\begin{table}[h]
\centering
\small
\begin{tabular}{p{0.30\linewidth} p{0.60\linewidth}}
\toprule
\textbf{Predicate} & \textbf{Meaning} \\
\midrule
\texttt{on(x, y)} & Block \texttt{x} is on block \texttt{y}. \\
\texttt{on-table(x)} & Block \texttt{x} is on the table. \\
\texttt{clear(x)} & Block \texttt{x} has no block on top of it. \\
\bottomrule
\end{tabular}
\caption{Predicates for SayCan.}
\label{tab:saycan-predicates}
\end{table}

Table\,\ref{tab:saycan-predicates} shows the predicates used in Saycan. An important difference between Saycan and Blocksworld is that SayCan can push blocks off the table, causing them to disappear.

\subsubsection{Existing Results}
In a previous version of \APPABBREV{}, we evaluated on SayCan. The capabilities learned in those runs are the ones reported in Fig.~\ref{tab:capabilities}. These models were learned using the same capability-learning code as the current version of \APPABBREV{}; however, both \APPABBREV{}-E and \APPABBREV{}-S have since been substantially improved. Therefore, we do not report variational distance results for the SayCan models. Instead, we include these learned capabilities only as examples of the types of capabilities that \APPABBREV{} can learn.

\section{Implementation Details and Setup}

In this section we discuss the implementation details and setup.

\subsection{MCQS Algorithm}
\label{sec:mcqs_alg}

\begin{algorithm}[H]
\caption{\textbf{Monte Carlo Query Search (MCQS)}}
\label{alg:mcqs}
\begin{algorithmic}[1]
\STATE \textbf{Inputs:} BBAI $\mathcal{A}$, simulator $\mathcal{S}_\mathcal{E}$, representation $\alpha : \mathcal{X} \rightarrow S$
\STATE \textbf{Output:} pessimistic capability model $h^\dagger$
\STATE $\mathcal{T}_0 \leftarrow random\_walk(\mathcal{S}_\mathcal{E})$ 
\STATE $C \leftarrow discover\_capabilities(\mathcal{T}_0, \alpha)$
\STATE $\mathcal{D} \leftarrow initialize\_dataset(C)$  
\STATE $h^\dagger_\land , h^\dagger_\lor  \leftarrow construct\_models(C,\mathcal{D})$
\STATE $x_i \leftarrow reset(\mathcal{S}_\mathcal{E})$
\WHILE{not $stop\_condition()$}
    \STATE $\xi\leftarrow set\_revisitation\_incentive()$
    \STATE $\pi \leftarrow synthesize\_query(\alpha(x_i), h^\dagger_\land , h^\dagger_\lor, \xi )$
    \STATE $\overline{\overline{x}} \leftarrow execute\_query(\mathcal{A}, \mathcal{S}_\mathcal{E}, x_i, \pi)$ 
    \STATE $\mathcal{D}\leftarrow update\_dataset(C,\mathcal{D}, \overline{\overline{x}}, \alpha)$
    \STATE $C\leftarrow update\_capabilities(C, \overline{\overline{x}}, \alpha)$
    \STATE $h^\dagger_\land , h^\dagger_\lor  \leftarrow update(h^\dagger_\land ,h^\dagger_\lor ,C,\mathcal{D})$
    \STATE $x_i \leftarrow sample\_initial\_state(\overline{\overline{x}}, \mathcal{S}_\mathcal{E})$
\ENDWHILE
\RETURN $h^\dagger_\land$
\end{algorithmic}
\end{algorithm}

In this section, we provide the full \APPABBREV{} algorithm, shown in Alg.\,\ref{alg:mcqs}. As mentioned in Section\,\ref{sec:ml_objective}, \APPABBREV{} takes as input only the BBAI $\mathcal{A}$, the simulator $\mathcal{S}_\mathcal{E}$ for environment $\mathcal{E}$, and the representation function $\alpha: \mathcal{X}\rightarrow S$ (line 1). It outputs a capability model $h^\dagger$ for the capabilities discovered and learned for $\mathcal{A}$ in $\mathcal{E}$ (line 2).

As discussed in Sec.\,\ref{sec:mcqs}, \APPABBREV{} begins by performing a random walk in $\mathcal{S}_\mathcal{E}$ to collect an initial set of low-level transitions $\mathcal{T}_0$, which are used to discover an initial capability set $C$ (lines 3--4). The process for constructing capabilities is described in Sec.\,\ref{sec:mcqs}. The algorithm then initializes the dataset of observed state-capability-state transitions $\mathcal{D}$ from $C$ and constructs the initial pessimistic and optimistic capability models, $h^\dagger_\land$ and $h^\dagger_\lor$, respectively (lines 5--6). The process for constructing $h^\dagger_\land$ and $h^\dagger_\lor$ is described in Sec.\,\ref{sec:hypothesis-space}.

Until the stop condition is triggered (line 8), \APPABBREV{} repeatedly synthesizes a query, executes it, and updates both the dataset and the learned hypotheses. In the first iteration, the query starts from the initial state of $\mathcal{S}_\mathcal{E}$ (line 7). In subsequent iterations, the initial state $x_i$ is chosen by sampling (line 15): if the previous query policy was predicted to be distinguishing and the maximum number of observed changes in the symbolic state space is less than 100, then a resulting state from the previous query is reused; otherwise, a non-sink state from $\mathcal{D}$ is sampled with probability inversely weighted by the number of times it has been observed. Details on initial-state sampling are provided in Appendix\,\ref{sec:init_state_sample}.

The state revisitation value $\xi$ is then dynamically set based on how recently new information has been collected (line 9); details are provided in Appendix\,\ref{sec:emp_add_info}. Given the initial symbolic state $s_i=\alpha(x_i)$, and the current models $h^\dagger_\land$ and $h^\dagger_\lor$, \APPABBREV{} constructs a distinguishing MDP $\mathcal{P}^\dagger(s_i,h^\dagger_\land,h^\dagger_\lor)$ (Def.\,\ref{def:dist_mdp}). This MDP is solved with $\xi$ by either \APPABBREV{}-S or \APPABBREV{}-E to produce a query policy $\pi:S\rightarrow C$ (line 10). For the random-query ablation used to evaluate \APPABBREV{} in Sec.\,\ref{sec:empirical_results}, a sequence of 30 capabilities is sampled instead.

The query policy is then executed 25 times from the initial low-level state $x_i$ by resetting $\mathcal{S}_\mathcal{E}$ to $x_i$ and having $\mathcal{A}$ follow the policy. At each symbolic state $s$, the next capability is selected as $c=\pi(s)$ and executed by $\mathcal{A}$. A rollout terminates if $\pi$ does not specify a capability for the current state $s$ or if more than 20 capabilities have been executed in the current query iteration. Thus, each query produces a set of trajectories, denoted $\overline{\overline{x}}$, for the capabilities executed during the query (line 11). This set is used to update both $\mathcal{D}$ and $C$ (lines 12--13). The process for converting trajectories into transitions is described in Sec.\,\ref{sec:optimizations}. After this update, $h^\dagger_\land$ and $h^\dagger_\lor$ are updated (line 14).

As mentioned in Sec.\,\ref{sec:empirical_results}, we output the pessimistic model $h^\dagger_\land$, since it provides a safer representation of the learned capabilities (line 16).

\subsection{Dynamic policy synthesis}
\label{sec:dynamic_policy}

Both \APPABBREV-E and \APPABBREV-S synthesize non-stationary policies rather than fixed capability sequences, allowing execution to adapt to observed outcomes.

In \APPABBREV-E, the returned policy contains the MCTS tree and tracks the current node. After a capability is selected, the current node advances to the corresponding child. Once the resulting state is observed, the policy performs a forward update from that node along the maximum-UCB continuation. During this pass, predicted state distributions are updated and impossible capabilities are pruned. The affected MCTS node statistics are then updated in a backward pass. If the maximum-UCB capability changes during this update, the newly selected branch is forward-updated and included in the backward pass.

\APPABBREV-S represents states and capabilities separately and therefore constructs a bipartite policy graph $\langle N,E\rangle$, where nodes correspond to capabilities and edges correspond to observed states. Let $n_0\in N$ denote the initial node. Let $f:N\rightarrow C$ label each node with a capability, and let $g:E\rightarrow S\times\mathbb{Z}^+$ label each edge with a state and its depth in the MCTS tree.

For each state node in the MCTS tree produced by \APPABBREV-S, let $s$ be the represented state at depth $d$, and let $n_0$ be the graph node corresponding to the capability executed to reach this state. The UCB-maximizing capability $c$ is selected and a new node $n_1$ is added to $N$ together with edge $e=(n_0,n_1)$, where $f(n_1)=c$ and $g(e)=(s,d)$. The algorithm then recursively proceeds to the child corresponding to $c$, carrying forward node $n_1$.

For a capability node, let $n_0$ denote the carried-forward node. All child state nodes are considered. For a state node labeled $(s,d)$, if there exists an edge $(n'_0,n'_1)$ with $g((n'_0,n'_1))=(s,d)$, then edge $(n_0,n'_1)$ is added. Otherwise, the algorithm recursively proceeds to that child while carrying forward $n_0$.

Let $n$ be the current node in the graph. The next capability selected by the policy is $f(n)$. After observing the next state $s'$, if there exists an edge $e=(n,n')$ such that $g(e)=(s',\ldots)$, then the current node is updated to $n'$.

\subsection{\APPABBREV{}-E Implementation}

To improve the efficiency of \APPABBREV{}-E, we introduce two optimizations.
First, to prevent the tree from containing redundant branches, we prune any newly generated node whose support set is identical to that of an existing node. This avoids exploring multiple nodes that are effectively equivalent.
Second, to reduce the frequency of distribution expansions and total-variation distance computations, we limit expansion to three child nodes when a node is first expanded, and again each time it is revisited.
For rollout, we do 3 random policy rollouts.

\subsection{Initial State Sampling}
\label{sec:init_state_sample}
To encourage exploration of under-explored regions of the observed state space $S_o\subseteq S$, we sample the initial state for query synthesis and execution according to observation counts in $\mathcal{D}$. For each state $s\in S_o$, let $N(s;\mathcal{D})=|\{\langle s,c,s'\rangle\in\mathcal{D}\}|$ denote the number of observed transitions from $s$. Let $n_{\max}=\max_{s\in S}N(s;\mathcal{D})$. We assign each state $s$ sampling weight $w_s=(n_{\max}+1-N(s;\mathcal{D}))/\sum_{\tilde{s}\in S_o}(n_{\max}+1-N(\tilde{s};\mathcal{D}))$.  

When we are sampling a new initial state from the outcome of the previous query, the process is in the same, but we use only the observed resulting states from each iteration of the query instead of $S_o$.
\subsection{Empirical Evaluation Information}
\label{sec:emp_add_info}

\begin{table*}[t]
    \centering
    \rowcolors{2}{gray!13}{white}
    \begin{tabular}{l c}
    \toprule
        \textbf{Description} & \textbf{Value} \\ 
        \midrule
        Number of runs per query & 25\\
        Environment state horizon & 100\\
        Warm Start Random Capability Walks & 0\\
        Max Capability Sequence & 20\\
        MCTS Exploration Constant & $\sqrt{2}$\\
        MCTS Iteration Count & 1000\\
        Random Policy if no distinguishing policy found & True\\
        Early Stop Condition & 25 Queries with no new information\\
        \bottomrule
    \end{tabular}
    \caption{Hyperparameters used for setting \APPABBREV-E and \APPABBREV-S}
    \label{tab:hyperparameters}
\end{table*}

For running \APPABBREV{}, we considered many hyperparameters. In Table\,\ref{tab:hyperparameters} we list all the additional hyperparameters we used when designing \APPABBREV{}.

For $\xi$, we scale it according to the number of queries since the last significant update. Let $n$ be the number of queries since the last observed information and let $\xi_m$ be the maximum exploration constant. Then $\xi$ is set as $\xi = \frac{n}{25}\xi_m$. Note 25 is the number of queries before we early stop, meaning $\xi$ can never exceed $\xi_m$.

Setting $\xi$ is an exploration--exploitation trade-off: if $\xi$ is too large, hard-to-reach capabilities may be starved because reducing the missing-effect probability below a threshold may require exponentially many samples. In this work, we set $\xi_m=10^{-5}$.

Finally, First responders contains many irreversible actions requiring execution from the initial state to reach specific states; therefore, for this domain, instead of initial-state sampling, we reset the simulator every 100 steps.

\section{Additional Results}
\label{appendix:add_results}

As discussed in Sec.~\ref {sec:results}, we evaluated \APPABBREV{} further by running an ablation where $\xi=0$ on 3 PDDLGym domains. We also evaluated models produced by \APPABBREV{}-S on there accuracy on reachability.

\subsection{No Revisitation Incentive}

\begin{figure*}
    \centering
    \includegraphics[width=1\linewidth]{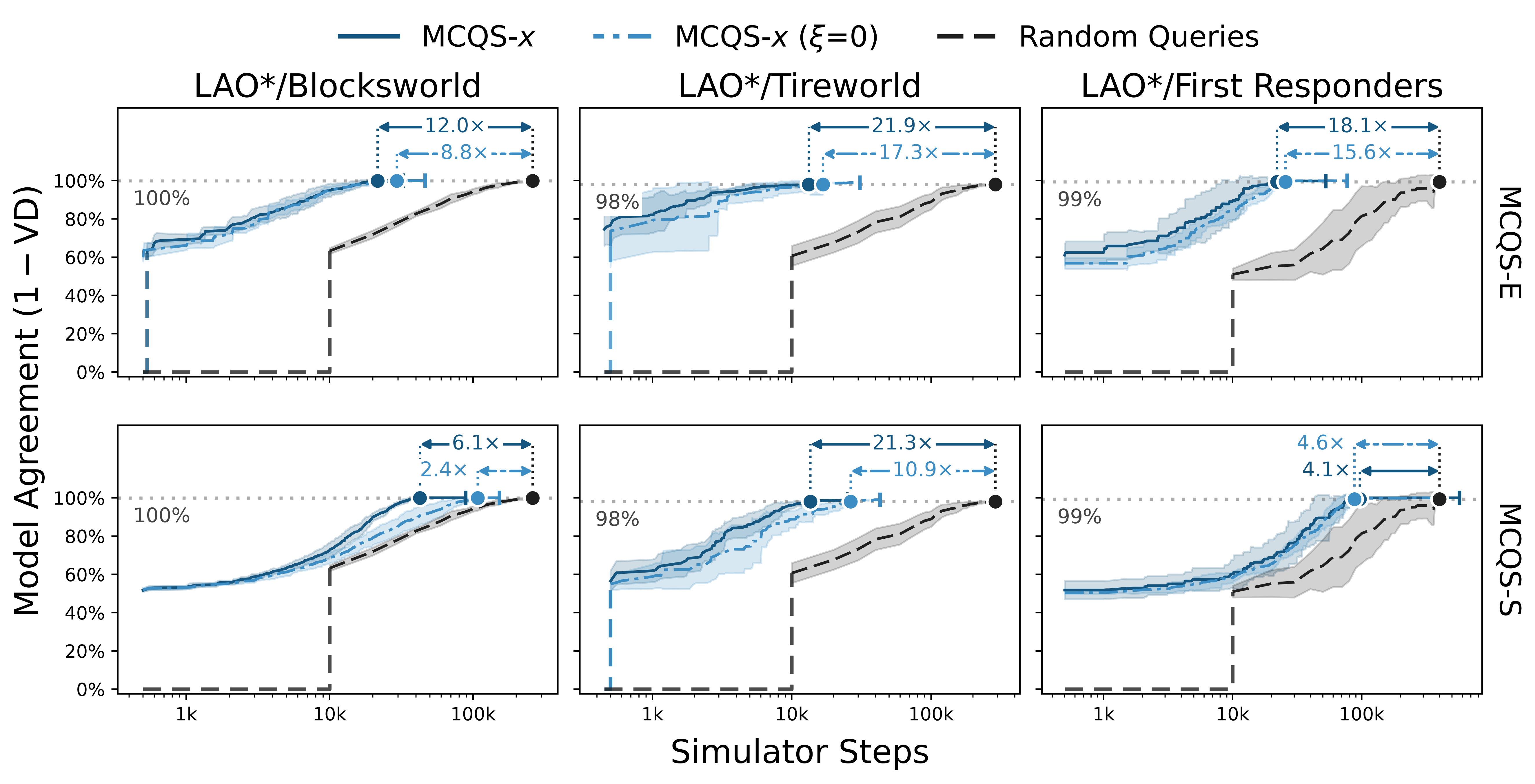}
    \caption{Model agreement (1 - weighted VD) of \APPABBREV{}-E and \APPABBREV{}-S on three evaluation problems. Shaded regions indicate one standard deviation over multiple runs. Annotated arrows indicate the simulator step-count ratio between each \APPABBREV{} approach and the Random Query ablation baseline.}
    \label{fig:xi_vd_results}
\end{figure*}

\begin{figure}
    \centering
    \includegraphics[width=\linewidth]{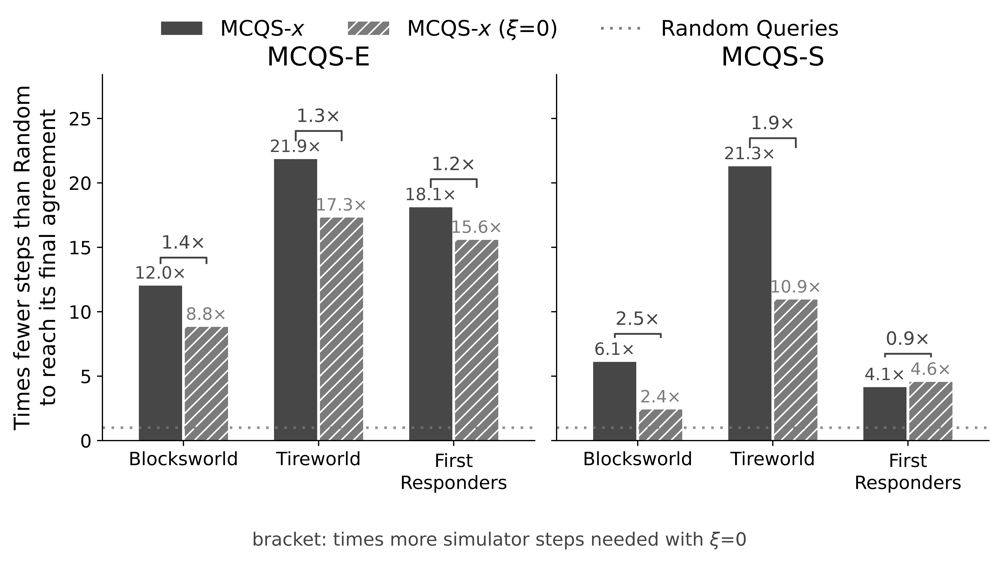}
    \caption{Times fewer simulator steps \APPABBREV{}-S and \APPABBREV{}-E had on three evaluation problems compared to random querying.}
    \label{fig:xi_sim_steps}
\end{figure}

We evaluated three PDDLGym domains (Blocksworld, Tireworld, and First Responders) using \APPABBREV{}-E and \APPABBREV{}-S with the parameters from the main paper and with the revisitation incentive disabled ($\xi=0$). We additionally evaluated Random Querying as a baseline. We report averages over five independent runs.

Fig.\,\ref{fig:xi_vd_results} shows that \APPABBREV{} learns accurate capability models both with and without the revisitation incentive. In most settings, \APPABBREV{} achieved at least 98\% model agreement with a non-zero $\xi$ with fewer simulator steps than with $\xi=0$, suggesting that the revisitation incentive can improve the quality of the policies tried. However, the effect is not uniform: \APPABBREV{}-S is more sensitive to $\xi$, and in some domains a non-zero $\xi$ requires substantially more simulator steps before stopping. Thus, while $\xi$ can improve sample quality, setting it too high may increase the total model learning cost.

This trade-off is highlighted by Fig.\,\ref{fig:xi_sim_steps}, which shows the number of simulator steps until the stopping condition is triggered. For \APPABBREV{}-E, using a non-zero $\xi$ required 20\% to 40\% more simulator steps than $\xi=0$. For \APPABBREV{}-S, the effect ranged from 10\% fewer steps to 150\% more steps, depending on the domain. Overall, $\xi$ can lead to better intermediate policies and faster agreement, but may also increase the total simulator steps needed before termination.

\subsection{Reachability Methodology}

To evaluate the reachability predictions of a model $h^\dagger$ learned by \APPABBREV{}-S, we uniformly sample an initial state $s_i$ from the observed states $S_{\mathrm{observed}}$. We then uniformly sample a capability $c$ that $h^\dagger$ predicts is achievable from $s_i$ and sample a possible successor state from $h^\dagger$. From this predicted state, we repeat the process until constructing a sequence of five capabilities. We reject sequences for which no valid capability is available before reaching length five.

We execute each sequence 100 times on the BBAI $\mathcal{A}$, resetting the simulator to $s_i$ before each execution. For each capability $c$ in the sequence, $\mathcal{A}$ attempts to achieve the intent associated with $c$. Execution continues until the sequence is completed or $h^\dagger$ classifies an observed transition as inconsistent with its prediction.

For each sequence, we compute the fraction of the 100 executions for which $h^\dagger$ remains consistent throughout the entire sequence. We report the mean and standard deviation of this accuracy across sampled sequences.

In Crafter, due to CSteve1 being set to greedy and evaluating a temporal length of 1, most capabilities were invalid resulting in only a handfull of valid 5-capability sequence. Therefore, we ran \APPABBREV{} on Crafter with a temporal length of 4 and set the agent to stochastic. While this creates more opportunities for the agent to achieve the intent, it also increases the difficulty of capability model learning due to a significant increase in stochasticity. However, as reported, \APPABBREV{}-S was able to learn a model that had a reachability accuracy of $78\%\pm 16\%$.

\section{QACE Baseline}
\label{app:qace}
 
In this section we describe how we ran QACE~\citep{verma23neurips}, the closest existing
approach to the problem addressed in this paper. We describe the inputs we supplied, the compilation
required to express our learning targets in QACE's model class, and the reasons QACE can be run only
on symbolic agents.
 
\subsection{Inputs and setup}
\label{app:qace-inputs}
 
QACE does not discover capabilities. It requires a predicate vocabulary $V$ and a list of
capabilities, each identified with an intent, and learns for each supplied capability a model with
conjunctive preconditions and probabilistic effects. We supplied the predicate vocabulary induced by
the representation function $\alpha$ used by MCQS, so that both methods
operate over the same symbolic state space. We supplied the capabilities discovered by MCQS
(Sec.~\ref{sec:mcqs}) as the capability list. QACE therefore begins each run with the output of the
discovery step, and the comparison in \S E1 measures model-learning efficiency alone.
 
\subsection{Compiling disjunctive conditions}
\label{app:qace-dnf}
 
MCQS represents a capability as a set of conditional-effect rules whose conditions are Boolean
formulas over $V$ (Def.~\ref{def:capability}), and the pessimistic and optimistic conditions
constructed are disjunctive. QACE supports only conjunctive preconditions,
so such capabilities are not expressible in its model class.
 
We therefore compile disjunctions away before passing capabilities to QACE. For each capability $c$
and each rule $r \in h_c$, we rewrite $\mathrm{cond}(r)$ in disjunctive normal form as
$\bigvee_{k=1}^{m} \varphi_k$, where each $\varphi_k$ is a conjunction of literals, and construct $m$
capabilities $c^{(1)}, \dots, c^{(m)}$. Each $c^{(k)}$ retains the intent and effect distribution of
$c$ and has precondition $\varphi_k$, and each is passed to QACE as a separate capability.
 
This compilation affects the comparison in three ways. First, the number of capabilities QACE
receives is $\sum_{c}\sum_{r \in h_c} m_{c,r}$ rather than $|C|$, and $m_{c,r}$ grows with the number
of condition blocks in the effect partition $\Phi_c$. Second, QACE learns each $c^{(k)}$
independently and does not represent the fact that these branches describe a single capability, so
the sample cost of learning $c$ is incurred $m$ times. Third, the disjunctive normal form is derived
from the model learned by MCQS, so the compilation supplies QACE with information about the structure
of the conditions. We use it because without it QACE's hypothesis class does not contain the true
model.
 
\subsection{Query execution on symbolic and simulator-based agents}
\label{app:qace-symbolic}
 
QACE poses queries at a specified symbolic state: a query fixes an initial symbolic state $s$ and
asks the agent to execute a capability, or a short sequence of capabilities, from $s$. Executing such
a query requires an inverse of the representation function, that is, a state $x \in X$ with
$\alpha(x) = s$ to which the simulator can be reset.
 
For symbolic agents this requirement is satisfied directly. For the LAO* agent on PDDLGym domains,
$\alpha$ is the identity, every symbolic state named in a query is instantiable, and the simulator
can be reset to it. All QACE runs reported in this paper use symbolic agents.
 
For BBAIs operating in non-symbolic simulators, including MiniGrid, Overcooked, Crafter, and SayCan,
$\alpha$ is many-to-one and has no inverse. A symbolic state such as
$\textit{agent-at}(\mathrm{NW}) \wedge \neg\textit{carrying-blue-key}()$ corresponds to a set of grid
configurations, and the simulator functionality, namely resetting
to an initial state, reverting to a previously encountered state $x \in X$, stepping with an action,
and querying available actions, does not include constructing or sampling a state from that set.
MCQS avoids this requirement, since a query is $\langle x_0, \pi, n \rangle$ where $x_0 \in X$ is an
environment state that has already been visited (Def.~\ref{def:query}) and the policy is executed
forward from $x_0$.
 
Executing a QACE query on such a simulator is possible in principle by searching for a reachable $x$
with $\alpha(x) = s$. This requires solving a reachability problem in $X$ for every query, with no
guarantee that the abstract state is reachable, and the cost of this search is charged to the same
simulator-step budget used for evaluation. We therefore report QACE on symbolic agents only.
 
\subsection{Scaling}
\label{app:qace-scaling}
 
The scaling of QACE on symbolic agents is limited by the grounded capability representation.
Capabilities are grounded (Sec.~\ref{sec:mcqs}), so MCQS constructs an intent for every
type-consistent grounding of each predicate observed in an effect, and $|C|$ grows as $O(|O|^{k})$
for a predicate of arity $k$. The compilation in Appendix~\ref{app:qace-dnf} increases this count
further. QACE generates queries over grounded state-capability pairs, so a grounded capability set
combined with a grounded predicate vocabulary causes the per-query search space to grow rapidly with
$|O|$.
 
QACE learned an accurate model for LAO* on Blocksworld, where the object set is small enough that the
grounded capability set remains small. For all other domains, including the remaining PDDLGym domains
whose agents are symbolic and whose queries are therefore executable, QACE produced no model within
50 hours. Table~\ref{tab:qace} reports the configuration and outcome for each domain.
 
\begin{table}[h]
\centering
\small
\begin{tabular}{lll}
\toprule
Domain & Agent  & Outcome \\
\midrule
Blocksworld             & LAO*                 & Model learned \\
First Responders        & LAO*                 & No model  \\
Tireworld               & LAO*                 & No model  \\
Prob. Elevators & LAO*                 & No model  \\
Depot                   & LAO*                 & No model  \\
\midrule
MiniGrid                & ReAct (GPT-5.4-nano) & Not runnable \\
Overcooked              & HDDLGym              & Not runnable \\
Crafter                 & CSteve1              & Not runnable \\
Crafter                 & Qwen CSteve          & Not runnable \\
SayCan                  & SayCan               & Not runnable \\
\bottomrule
\end{tabular}
\caption{configuration and outcome of QACE runs for each domain. $|C|$ is the number of capabilities
discovered by MCQS and supplied to QACE, and $|C_{\mathrm{DNF}}|$ is the number after compiling
disjunctive conditions into conjunctive capabilities (Appendix~\ref{app:qace-dnf}). Not runnable
indicates that $\alpha$ is not invertible for the agent's simulator
(Appendix~\ref{app:qace-symbolic}).}
\label{tab:qace}
\end{table}

\section{Limitations}
\label{sec:limitations}

A key result of this work is that accurately modeling BBAI capabilities requires an expressive grounded representation; lifting and aggressive generalization can produce inaccurate capability models. The consequence is that \APPABBREV{} must learn grounded capabilities directly from observed transitions, making model quality dependent on transition coverage. In practice, this requires tracking all observed transitions between symbolic states and capabilities. Although this space is substantially smaller than the underlying environment state space, for symbolic states $S$ and capabilities $C$, the resulting worst-case space and time complexity is $O(S^2C)$. While this level of expressivity is currently necessary for accurate capability learning, future work is needed to develop more efficient representations that preserve modeling fidelity while improving scalability.

A related limitation is that BBAIs and their environments are often highly stochastic. This substantially increases sample complexity because repeated policy executions are required both to observe rare effects and to accurately estimate their probabilities.

\section{Computational resources}
\label{sec:comp_resources}

Experiments were conducted on two hardware configurations: (1) an Intel Core i9-9900 CPU @ 3.10GHz with an NVIDIA GeForce RTX 2080 and 64GB RAM, and (2) an AMD Ryzen Threadripper PRO 7975WX (32 cores) with an NVIDIA RTX 6000 Ada Generation GPU and 64GB RAM.

MiniGrid, ReAct, Overcooked, and the PDDLGym domains were run on the i9 system using 15 parallel runs for \APPABBREV{}-E, \APPABBREV{}-S, and the Random Query baseline. Smaller PDDLGym domains, such as blocksworld and tireworld, required less than one hour each. First responders required approximately 3 hours, while probabilistic elevators and ReAct each required approximately 24 hours. Constructing the evaluation datasets required roughly one additional day. In total, this portion of the experiments required approximately 60 GPU-hours on hardware comparable to the i9 setup.

Both Crafter variants required approximately 24 hours each on the Threadripper system, with an additional 8 hours each for dataset construction. Running the SayCan experiments required approximately 72 hours per run, with only two runs executable in parallel, resulting in roughly 563 total hours. Overall, the Crafter and SayCan experiments required approximately 625 hours on the Threadripper system.

\section{Broader impacts}
\label{sec:broader_impact}

This work studies methods for learning interpretable capability models of black-box AI agents. The primary goal is improving transparency, reliability, and safety assessment by identifying the conditions under which agents succeed, fail, or exhibit unintended side effects. Such capability models may help users deploy AI systems more safely and help developers identify behavioral limitations.

We do not foresee direct harmful applications of this work. However, like many evaluation and interpretability techniques, these methods could potentially be used to characterize weaknesses or behavioral patterns of deployed agents. Overall, we believe the primary impact of this work is enabling more reliable understanding and evaluation of increasingly complex AI systems.

\end{document}